\documentclass[10pt,twocolumn,letterpaper]{article}

\usepackage[pagenumbers]{cvpr} %

\definecolor{mygreen}{HTML}{4CAF50}
\usepackage[accsupp]{axessibility}  %
\definecolor{cvprblue}{rgb}{0.21,0.49,0.74}
\usepackage[pagebackref,breaklinks,colorlinks,allcolors=cvprblue]{hyperref}
\usepackage{makecell}
\usepackage[most]{tcolorbox}

\usepackage{bbm}

\def\paperID{*****} %

\def\confName{CVPR}
\def\confYear{2026}

\title{%
Same Content, Different Answers: Cross-Modal
Inconsistency in MLLMs}

\author{
  {
  Angela van Sprang\textsuperscript{1}\footnotemark[1],
  Laurens Samson\textsuperscript{1,2}\footnotemark[1]},\\
  Ana Lucic\textsuperscript{1},
  Erman Acar\textsuperscript{1},
  Sennay Ghebreab\textsuperscript{1},
  Yuki M. Asano\textsuperscript{3}\\
  \textsuperscript{1}University of Amsterdam \hspace{.8cm} 
  \textsuperscript{2}City of Amsterdam \hspace{.8cm} 
  \textsuperscript{3}University of Technology Nuremberg\\
}

\begin{document}
\maketitle
\footnotetext[1]{Shared first author. \texttt{\{a.v.vansprang,l.samson\}@uva.nl}}
\begin{abstract}
Multimodal large language models (MLLMs) are trained to represent vision and language in a shared space. But does this joint representation enable consistent reasoning across modalities? We introduce \textbf{REST} and \textbf{REST+} (Render-Equivalence Stress Tests), two benchmarks for systematically evaluating cross-modal consistency. Each sample presents \textbf{semantically identical information} in three forms (image, text, and mixed), allowing us to measure whether models produce consistent outputs regardless of modality.
Evaluating 15 state-of-the-art MLLMs, we find that none reason consistently across modalities, with substantial variation in the degree of inconsistency. Neither rendering text as images nor images as text resolves this problem, even when controlling for OCR errors. We further show that visual characteristics (color, resolution, but not font) and the number of vision tokens affect performance even when text is correctly recognized. Finally, our consistency score correlates with the cross-modal cosine similarity in embedding space, suggesting a mechanistic explanation: inconsistent reasoning arises when text and image representations occupy distinct regions of the joint space. Data and code are available on \url{https://github.com/angelavansprang/Same-Content-Different-Answers}.
\end{abstract}

\section{Introduction}

Multimodal large language models (MLLMs)~\cite{achiam2023gpt, alayrac2022flamingo, yin2024survey, liu2023visual, liu2024improved, dai2023instructblip} have achieved remarkable progress across diverse multimodal tasks, demonstrating strong capabilities in visual question answering~\cite{yue2024mmmu, yue2024mmmupro, liu2023mmbench}, document understanding~\cite{mathew2021docvqa, masry2022chartqa}, and complex reasoning~\cite{radford2021learning, alayrac2022flamingo, liu2023visual}. 
The success of MLLMs suggests they have learned to integrate visual and textual information for multimodal understanding seamlessly, even though these modalities might not explicitly be integrated. In particular, recent work points out the existence of a \textit{modality gap} in different MLLMs, where text and image embeddings occupy different regions in the joint embedding space \cite{liang2022mind, shukor2024implicit, papadimitriou2025interpreting}. The smaller this gap, the better the downstream task performance \cite{shukor2024implicit, eslami2024mitigategapinvestigatingapproaches}. 

Simultaneously, recent work by the authors of DeepSeek-OCR~\cite{wei2025deepseek} explores the idea of rendering text as images to reduce token costs. They demonstrate that compressing 10 text tokens into a single visual token maintains 97\% optical character recognition (OCR) accuracy, offering substantial computational savings for text-heavy inputs. However, this raises a fundamental question given the modality gap: \textit{when a model successfully reads text from an image, does it reason about that information as effectively as when receiving native text}? If not, this could pose critical problems: MLLMs might perform better on the text modality (given that their foundation is an LLM). They might produce incorrect answers based solely on the prompt modality, rather than on the information it contains. We refer to this as \textbf{cross-modal inconsistency}: obtaining inconsistent results on prompts with semantically identical (``same content'') information in different modalities. %

Existing work has documented cross-modal inconsistencies in MLLMs \cite{yue2024mmmupro, shu2025large, alonso2025vision, sim2025can, samson2024privacy}, but prior benchmarks either evaluate only a single model~\cite{zhang2024crossmodalconsistencymultimodallarge} or do not control for actual readability (OCR, \citep{chen2024omnixr}), potentially conflating text recognition failures with reasoning inconsistencies.

\begin{figure*}
    \centering
    \includegraphics[width=\linewidth]{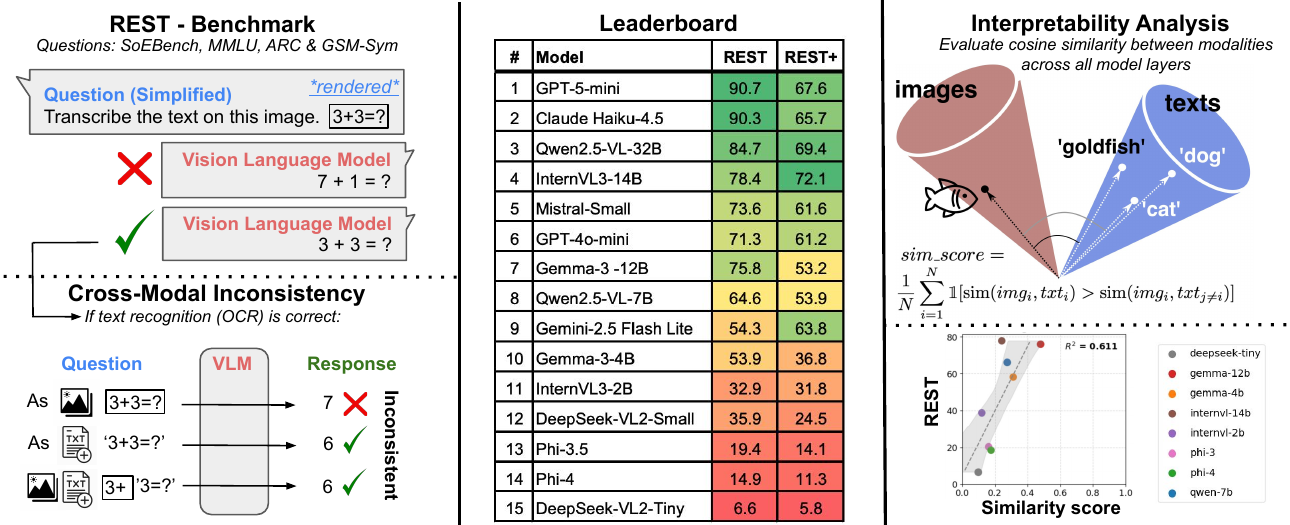}
    \caption{\textbf{Summary of our work}. 
    \textit{Left:} Our \textbf{REST} benchmark measures whether MLLMs can consistently reason over identical information across modalities. We first verify text recognition (OCR) capability, then evaluate the same question in three modalities (text, image, mixed). Cross-modal inconsistency occurs when models produce different answers depending on the input format.
     \textit{Center:} RER consistency score measures the degree to which a model outputs the same answer in all modalities. We evaluate 15 MLLMs (on OCR correct, sorted on \textbf{REST}) and find that the degree of cross-modal inconsistency varies substantially across models even when controlling for OCR. 
    \textit{Right:} Matching samples (i.e., different modalities containing the same information) show higher cosine similarity than non-matching ones, and the extent of this difference correlates with the consistency score on our benchmark.}
    \label{fig:placeholder}
\end{figure*}
To address this gap, we introduce \textbf{REST} (Render-Equivalence Stress Tests) and \textbf{REST+} to systematically evaluate cross-modal reasoning consistency under controlled OCR difficulty. 
Both benchmarks present the same semantic content in three formats: text-only, image-only, and mixed (question in text, context in image). 
We use three established benchmarks where we render text as images (MMLU~\cite{hendrycks2020mmlu}, ARC~\cite{clark2018arc}, GSM-Symbolic~\cite{mirzadeh2024gsm}) and \textsc{SoEBench}, our newly generated system-of-equations task with easy-to-recognise symbols. 
\textsc{SoEBench} ensures minimal OCR burden and guarantees zero prior exposure, reducing potential memorisation by MLLMs. 
All analyses are performed conditional on the model performing OCR correctly for the given question.
\textbf{REST+} extends \textbf{REST} by permuting each image in 10 different ways, varying the font, resolution, and text color.
This allows us to evaluate the consistency of reasoning skills across formats, including the amount of visual tokens. 
Our research questions are:
\begin{enumerate}[\noindent {}]
    \item \textbf{RQ1}: Do current frontier MLLMs exhibit cross-modal inconsistency? If so, which modalities perform best? %

    \item \textbf{RQ2}: Is cross-modal inconsistency due to OCR performance rather than true modality differences?
    \item \textbf{RQ3}: Do visual characteristics such as resolution (\textbf{RQ3a}) or font and text colour (\textbf{RQ3b}) influence cross-modal inconsistency, and if so, how?

    \item \textbf{RQ4}: Given the internal representations of the same concept in different modalities, is there a correlation between the cross-modal similarity of internal representations and the degree of cross-modal inconsistency?
 
\end{enumerate}

We have three main contributions. 
First, we introduce \textbf{REST} and  \textbf{REST}+ as benchmarks to measure cross-modal inconsistency. Unlike previous benchmarks, we include a new set of tasks (\textsc{SoEBench}) that is guaranteed to not be seen during pre-training and control for OCR complexity. 
Next, we evaluate 15 frontier MLLMs and find substantial inconsistencies across modalities (at least $\sim$10\% inconsistency), even when controlling for OCR. This leaves a notable gap of solvable questions that current models fail to capture.
Finally, we analyse the internal representations of matching samples (i.e., samples with the same information in different modalities) and find that they show higher cosine similarity than non-matching pairs, and that this similarity magnitude correlates with our consistency score.

\section{Our Benchmarks: REST and REST+}

\paragraph{REST} 
Our benchmark consists of four evaluation tasks:
\begin{itemize}
    \item \textbf{OCR}: Extract rendered text from the image.
    \item \textbf{Text}: Answer textual questions.
    \item \textbf{Image}: Answer questions rendered as images.
    \item \textbf{Mixed}: Answer questions with the context as image and the question as text (or, in case of multiple-choice questions, the question is the image and the options are text). Note that this is not per se a neutral midpoint between text and image and it can be flipped by swapping text and image components. We find no configuration to be inherently easier across MLLMs (see App. \ref{app:extended}).
\end{itemize}
\noindent The data consists of our \textsc{SoEBench} and three established benchmarks: MMLU~\cite{hendrycks2020mmlu}, ARC~\cite{clark2018arc}, and GSM-Symbolic~\cite{mirzadeh2024gsm}. To isolate cross-modal consistency from OCR failures, we minimise OCR complexity (i.e. we prevent models from failing to understand the images) by excluding questions with more than 800 characters (threshold decided by initial experiments) and questions with LaTeX code (for which OCR is ambiguous). 
We render images with white backgrounds and high-resolution (DPI 200) black DejaVu Sans text.

\paragraph{REST+}
We introduce \textbf{REST+} to study the effect of visual characteristics (such as resolution, font and colour) on cross-modal inconsistency. \textbf{REST+} increases complexity by creating 10 visual permutations of each image while preserving semantic content. For computational feasibility, we only include the text and image tasks.
We use MMLU with the same question filtering as in \textbf{REST}.
We create 10 permutations for each question: 9 combinations from 3 font families (DejaVu Sans, Courier New, Cursive), 3 DPI values (50, 100, 200), and one coloured variant (red, green, blue, cyan, magenta, or yellow) at 200 DPI using DejaVu Sans. We manually verified that the text is legible at 50 DPI for each font. We randomly sample 10\% per subject class for computational feasibility, which results in 1085 final questions. Sample images are provided in the Appendix. 

\paragraph{\textsc{SoEBench}}
is a task suite based on solving systems of linear equations presented in text, image, or mixed modalities included in \textbf{REST} and \textbf{REST+}. Each question contains symbolic variables and requires basic algebraic reasoning, with all instances generated to have a single integer solution.
Its restricted symbol set (digits 0–9 and letters A–E) reduces OCR complexity, ensuring that performance differences reflect reasoning rather than recognition errors. 
Since it is newly generated, we can guarantee existing MLLMs have had no prior exposure to it.

\section{Benchmarking Methodology}

For each \textbf{REST} question, the MLLM receives four input prompts. First, we verify OCR performance by letting the model transcribe the image, ensuring that the model can read the question. For the remaining tasks (text, mixed, image), we employ Chain-of-Thought (CoT) prompting. 
All prompts are shown in the Appendix.
We limit the output to 1024 tokens for all tasks, except for \textsc{SoEBench}, for which we allow 2048 output tokens because 1024 is not always sufficient. We set the temperature to 0 for reproducibility. We parse answers using regular expressions and treat format-invalid responses as incorrect.

We introduce three metrics to evaluate our benchmark on cross-modal inconsistency. 
We measure answer consistency across modalities with the \textit{Render-Equivalence Rate} (RER, similar to the consistency score from \citet{zhang2024crossmodalconsistencymultimodallarge}):
\setlength{\abovedisplayskip}{2pt}
\begin{multline}
\text{RER} = \frac{ |\{ x \mid f(x_{\text{t}}) = f(x_{i}, z_{i}) = f(x_{m}, z_{m}) \}| }{N} ,
\end{multline}
\setlength{\belowdisplayskip}{2pt}

where $x$ denotes text prompts and $z$ denotes images for input type (text, image or mixed), $f$ is the model, and $N$ is the dataset size. This metric represents the fraction of questions for which the model gives the same answer for all modalities, where 1 denotes a perfectly consistent model.

We also evaluate inconsistency in questions solvable only through specific modalities, i.e., those answerable in at least one, but not all, modalities.  We use the \textit{Cross-Modality Failure Rate (CFR)}:
\begin{equation}
\text{CFR} = \frac{|\{ q \mid 1  \leq \sum_{m \in M} C(q,m) < |M|\}|}{N_{c}} ,
\end{equation}
where $N_c$ is the number of questions that are answered correctly in at least one modality, $M$ = \{text, image, mixed\}, and $C(q,m)$ indicates correctness (1 correct, 0 incorrect) for question $q$ and modality $m$. We exclude questions that failed for all modalities since these represent inability rather than inconsistency. A score of 0 indicates a perfectly consistent model.

We assess overall capability using \textit{Max Modal Coverage} (MMC):
\begin{equation}
\text{MMC} = \frac{|\{q \mid 1 \leq \sum_{m \in M} C(q,m)\}|}{N} ,
\end{equation}
which represents the fraction of questions solved through at least one modality. A score of 1 indicates (potentially) perfect performance on all questions, provided that each question is presented in the suitable modality.

To evaluate OCR performance, we use \textit{Character Error Rate} (CER), which measures character insertions, deletions, and substitutions and normalises by reference length. For the evaluation of cross-modality consistency, we include only the questions where text is perfectly recognised. We only evaluate letters and digits to focus on semantic recognition rather than punctuation.\footnote{
When extending the evaluation to include mathematical symbols we find negligible differences across datasets. Crucially, \textit{consistency scores remain unchanged} across all datasets.}

\section{Benchmarking Results: REST}
\label{ch:results}

\begin{table}[htb]

\centering
\footnotesize 
\caption{\textbf{REST scores for 15 MLLMs show varying degrees of cross-modal inconsistency.} RER and CFR scores are given for all questions and for those where models correctly recognised the rendered text (OCR\checkmark).}
\setlength{\tabcolsep}{0.45em}
\begin{tabular}{l|cc|ccc}
\toprule
\textbf{Model} 
& \multicolumn{2}{c|}{\makecell{\textbf{REST} (OCR\checkmark)}} 
& \multicolumn{3}{c}{\makecell{\textbf{REST} (All Questions)}} \\
\cmidrule(lr){2-3} \cmidrule(lr){4-6}
 & RER $\uparrow$ & CFR $\downarrow$ & RER $\uparrow$ & CFR $\downarrow$ & OCR\checkmark $\uparrow$  \\
 \midrule
Deepseek-Tiny~\cite{wu2024deepseek} & 6.6 & 98.0 & 6.5 & 98.1 & 70.3 \\
Phi-4~\cite{abdin2024phi} & 14.9 & 82.3 & 14.5 & 82.8 & 94.0 \\
Phi-3.5~\cite{abdin2024phi3} & 19.4 & 79.3 & 19.3 & 79.5 & 90.3 \\
InternVL3 (2B)~\cite{zhu2025internvl3} & 32.9 & 63.7 & 32.6 & 63.9 & 90.7 \\
Deepseek-Small~\cite{wu2024deepseek} & 35.9 & 60.9 & 35.9 & 60.9 & 97.9 \\
Gemma-3 (4B)~\cite{team2025gemma} & 53.9 & 42.3 & 52.3 & 44.0 & 83.5 \\
Gemini-2.5 Fl. Lite~\cite{team2023gemini} & 54.3 & 40.3 & 54.1 & 40.4 & 98.3 \\
Qwen-2.5 (7B)~\cite{bai2025qwen2} & 64.6 & 31.7 & 64.3 & 31.9 & 98.8 \\
GPT-4o-mini~\cite{hurst2024gpt} & 71.3 & 26.0 & 71.2 & 26.1 & 98.8 \\
Mistral-Small~\cite{mistral2025small} & 73.6 & 23.9 & 73.4 & 24.1 & 98.4 \\
Gemma-3 (12B)~\cite{team2025gemma} & 75.8 & 21.3 & 75.5 & 21.6 & 96.5 \\
InternVL3 (14B)~\cite{zhu2025internvl3} & 78.4 & 19.6 & 78.1 & 19.9 & 95.3 \\
Qwen-2.5 (32B)~\cite{bai2025qwen2} & 84.7 & 13.6 & 84.5 & 13.7 & 97.5 \\
Haiku-4.5 (Claude)~\cite{anthropic_claude_haiku_4_5} & 90.3 & 8.9 & 90.1 & 9.1 & 98.2 \\
GPT-5-mini ~\cite{openai_gpt_5_mini}~& \textbf{90.7} & \textbf{8.7} & \textbf{90.7} &\textbf{ 8.8} & \textbf{99.0} \\
\bottomrule
\end{tabular}
\label{tab:rest_consistency}
\end{table}

\subsection{RQ1: Cross-modal inconsistency in MLLMs}
Table \ref{tab:rest_consistency} presents RER and CFR scores for 15 state-of-the-art MLLMs, with separate scores for all data and only those questions with correct OCR ($OCR\checkmark$).
The scores are averaged over all four \textbf{REST} benchmarks.
The RER consistency scores vary substantially across MLLMs, ranging from 6.6\% to 90.7\%. Notably, RER scores are equal or slightly higher for all models when only evaluating questions with correct OCR. This indicates that OCR errors do affect consistency, and our efforts to control OCR complexity are justified.
Moreover, the model rankings are not affected by excluding incorrect OCR questions. Overall, closed-source models (GPT-5-mini~\cite{openai_gpt_5_mini} and Claude Haiku 4.5~\cite{anthropic_claude_haiku_4_5}) achieve the highest consistency scores (90.7\% and 90.3\%, respectively, when OCR is correct). 
Gemini 2.5 Flash Lite~\cite{team2023gemini} and GPT-4o-mini~\cite{team2023gemini} perform worse than other closed-source models. Among open-source models, Qwen-2.5 32B~\cite{bai2025qwen2} leads with an RER of 84.7\%.

The CFR results reveal concerning patterns: even GPT-5-mini fails to consistently solve 8.7\% of questions across all modalities despite solving them in at least one modality. This inconsistency even reaches 82.3\% for Phi-4~\cite{abdin2024phi}, which means that obtaining a correct answer depends on how the input is formatted (text, mixed, or image).

\begin{figure*}
    \centering
    \includegraphics[width=0.99\textwidth, trim=0 12 0 0, clip]{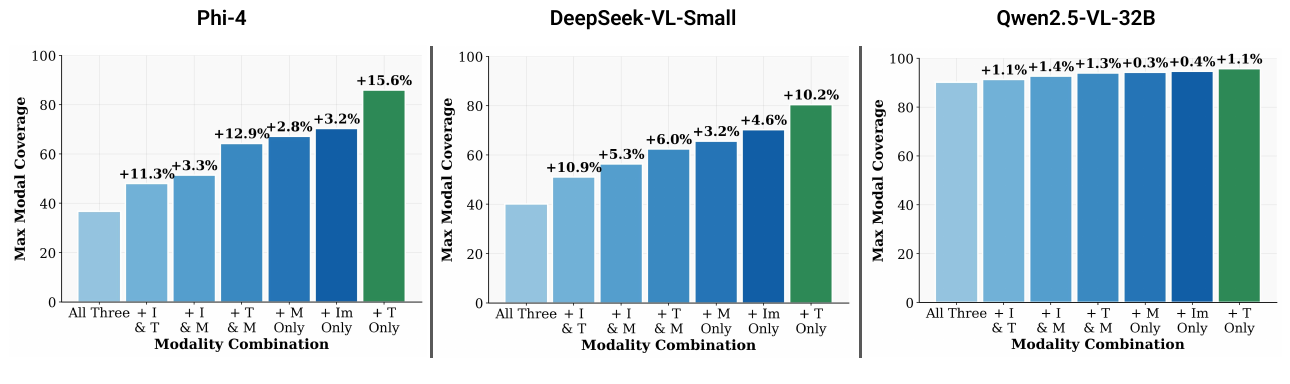}
    \caption{\textbf{Cross-modal inconsistency leaves model potential untapped.}
    This figure shows the cumulative distribution of correctly solved questions across sets of modalities (OCR-correct subset). 
    From left to right, the bars represent: the percentage of questions that can be solved in all three modalities, followed by including questions that can only be solved in fewer modalities, ending with the Max Modal Coverage (\textcolor{mygreen}{green}), which shows the percentage of questions that can be solved in at least one modality.
    Results are shown for GSM8k-Symbolic, which contains open-ended questions that cannot be solved by guessing.}
    \label{fig:max_modal_coverage}
\end{figure*}

\paragraph{Max modal coverage}
Given the presence of cross-modal inconsistency in MLLMs, we examine which modalities obtain more correct answers. Figure \ref{fig:max_modal_coverage} shows models with low, medium, and high consistency scores for GSM8k-Symbolic~\cite{mirzadeh2024gsm}. 
We report scores on this dataset because it contains open-ended questions that cannot be solved by guessing. 
The results show that models with low consistency could increase model performance drastically when solving questions in the optimal modality. In particular, MMC (i.e., fraction of questions that are solvable through at least one modality) for Phi-4 is 85.9\%, but only 36.7\% of questions can be solved in all modalities. So, 49.2\% of questions might obtain incorrect answers if they are not in the optimal modality. Similarly, DeepSeek-VL-Small~\cite{wu2024deepseek} can only solve 10.2\% of the questions through text alone. Qwen2.5-VL-32B shows the best consistency, where almost all questions can be solved in all modalities.

\begin{table}[t]
\centering
\footnotesize 
\caption{\textbf{Preference for Text.} This table presents accuracies for the text, mixed, and image tasks on MMLU, AI2-ARC, and GSM8k-Symbolic (OCR-correct subset). Bold values indicate the best-performing modality for each model-dataset combination. 
}
\setlength{\tabcolsep}{0.35em}
\begin{tabular}{l|ccc|ccc|ccc}
\toprule
\textbf{Model} 
& \multicolumn{3}{c|}{\textbf{MMLU~\cite{hendrycks2020mmlu}} $\uparrow$} 
& \multicolumn{3}{c|}{\textbf{AI2-ARC~\cite{clark2018arc}} $\uparrow$} 
& \multicolumn{3}{c}{\textbf{GSM-Sym~\cite{mirzadeh2024gsm}} $\uparrow$} \\
\cmidrule(lr){2-4} \cmidrule(lr){5-7} \cmidrule(lr){8-10}
 & Text & Mix & Img & Text & Mix & Img & Text & Mix & Img \\
 \midrule
Deepseek-T & \textbf{29.5} & 26.6 & 25.3 & \textbf{34.2 }& 28.7 & 27.5 & \textbf{15.5} & 0.8 & 0.7 \\
Deepseek-S & 54.9 & 54.9 & 51.2 & 69.1 & 68.7 & \textbf{70.0} & \textbf{67.3 }& 54.7 & 60.9 \\
GPT-4o-mini & \textbf{84.4 }& 77.2 & 77.0 & \textbf{93.6} & 89.3 & 89.9 & \textbf{91.4} & 86.7 & 89.0 \\
GPT-5-mini & \textbf{89.3} & 86.5 & 87.8 & \textbf{94.6 }& 92.7 & 93.3 & \textbf{95.1} & 93.7 & 94.1 \\
Gemini-2.5 FL & \textbf{81.6} & 79.7 & 78.4 & \textbf{90.4} & 89.0 & 88.8 & \textbf{79.4 }& 60.1 & 50.5 \\
Gemma-3-4B & \textbf{68.0} & 64.7 & 57.2 & \textbf{79.2} & 78.2 & 71.4 & \textbf{82.2} & 70.5 & 67.4 \\
Gemma-3-12B & \textbf{77.6} & 76.7 & 72.2 & \textbf{90.5} & 90.2 & 88.6 & \textbf{91.3} & 87.7 & 88.1 \\
Haiku-4.5 & \textbf{90.1} & 87.3 & 85.1 & 93.3 & 93.3 & 92.7 & \textbf{95.6} & 94.9 & 94.1 \\
InternVL3-2B & \textbf{61.6} & 59.0 & 56.6 & \textbf{76.0 }& 74.0 & 74.0 & 52.1 & 40.3 & \textbf{53.1} \\
InternVL3-14B & 82.9 & 81.4 & \textbf{83.0} & 92.9 & 92.9 & 94.2 & 91.2 & 91.9 & \textbf{93.0} \\
Mistral-Small & \textbf{84.1 }& 78.9 & 82.5 & \textbf{92.7} & 90.0 & 92.6 & 92.6 & 91.7 & \textbf{93.3} \\
Phi-3.5 & \textbf{61.1} & 41.5 & 32.5 & \textbf{73.3} & 54.3 & 37.9 & \textbf{66.9} & 42.9 & 48.3 \\
Phi-4 & \textbf{47.0} & 44.2 & 31.4 & \textbf{59.9} & 48.5 & 35.0 & \textbf{76.5 }& 55.8 & 54.5 \\
Qwen-2.5-7B & 72.6 & 71.8 & \textbf{72.9} & 86.1 & 87.0 & \textbf{88.3} & \textbf{84.4} & 82.0 & 82.1 \\
Qwen-2.5-32B & 83.3 & \textbf{84.1} & 83.5 & 92.9 & 92.9 & 92.9 &\textbf{ 93.6} & 93.1 & 93.0 \\
\bottomrule
\end{tabular}
\label{tab:established_benches}
\end{table}

\paragraph{Preference for text modality}
Using the different tasks in \textbf{REST}, we find that almost all models perform better in the text modality (see Table \ref{tab:established_benches} for the accuracies on MMLU, AI2-ARC and GSM8k-Symbolic). 
With a statistical t-test, and combining the answers from all benchmarks, we find that models prefer text over image ($t=17.7$, $p < 0.05$) and that image is more difficult than mixed ($t=-7.2$, $p < 0.05$). Several model families (Phi, Gemma, ChatGPT, and Claude) consistently achieve their best performance in text rather than other modalities (e.g., GPT-4o-mini obtains more than 7\% higher accuracy in text on MMLU). This raises the question whether we observe data contamination, where the questions of these text benchmarks were included in the models' training data, or whether the text modality is inherently more effective in MLLMs.

\subsection{RQ2: Controlling for OCR}
Therefore, we evaluate cross-modal inconsistency on data that all models have not seen before, while keeping OCR simple, by using our \textsc{SoEBench}.
Table \ref{tab:soebench} shows the consistency scores, task accuracies and OCR accuracies on this benchmark. Firstly, all models can solve OCR (near) perfectly, with the exception of DeepSeek-Tiny, which reproduces the example answer from the input prompt. Moreover, the Gemma and GPT models still outperform on the text modality. Surprisingly, Phi-4 now performs better on the image modality, while the other benchmarks showed higher performance on the text modality by a large margin.

We argue that OCR and data contamination are not the main reasons for cross-modal inconsistency, as models inherently perform better through the text modality.

\begin{table}[t]
\centering
\footnotesize 
\caption{\textbf{Models inherently perform better through the text modality.} These results on our \textsc{SoEBench} include consistency scores, accuracies for the different tasks and OCR accuracy (OCR-correct subset). Even when the data is novel and models achieve (nearly) perfect OCR scores, the text modality is still preferred. 
}
\setlength{\tabcolsep}{0.45em}
\begin{tabular}{l|c|ccc|c}
\toprule
\textbf{Model} & RER $\uparrow$ & Text $\uparrow$ & Mixed $\uparrow$ & Image $\uparrow$ & OCR $\uparrow$\\
\midrule
Phi-3.5 & 0.0 & 1.3 & 1.3 & 0.7 & 100.0 \\
Deepseek-Tiny & 0.0 & \textbf{0.7} & 0.0 & 0.0 & 0.0 \\
Deepseek-Small & 0.7 & \textbf{8.0} & 7.3 & 4.7 & 100.0 \\
Phi-4 & 1.3 & 13.3 & 13.3 & \textbf{17.3} & 100.0 \\
InternVL3 (2B) & 4.7 & 14.0 & 14.7 & \textbf{33.3} & 100.0 \\
Gemini-2.5 Flash Lite & 13.3 & 42.7 & 34.7 & \textbf{49.3} & 100.0 \\
Qwen-2.5 (7B) & 24.7 & 48.0 & 46.0 & \textbf{48.7} & 100.0 \\
GPT-4o-mini & 39.3 & \textbf{65.3} & 61.3 & 62.0 & 100.0 \\
Mistral-Small & 41.3 & 60.7 & 62.0 & 62.0 & 100.0 \\
Gemma-3 (4B) & 41.9 & \textbf{75.2 }& 58.9 & 57.4 & 86.0 \\
InternVL3 (14B) & 50.7 & \textbf{73.3} & 70.0 & 72.7 & 100.0 \\
Gemma-3 (12B) & 63.3 & \textbf{84.0} & 72.7 & 76.0 & 100.0 \\
Qwen-2.5 (32B) & 69.1 & 87.2 & \textbf{87.9} & 75.2 & 99.3 \\
GPT-5-mini & 91.9 & \textbf{98.7} & 95.3 & 97.3 & 99.3 \\
Haiku-4.5 (Claude) & 92.0 & 96.7 & 95.3 & \textbf{97.3} & 100.0 \\
\bottomrule
\end{tabular}
\label{tab:soebench}
\end{table}

\begin{table}[htb]
\centering
\footnotesize 
\caption{\textbf{`OCR first, then solve' strategy can hurt performance.} We show the difference in accuracy ($\Delta$) when MLLMs are instructed to first transcribe text from images before solving the task (OCR-correct subset). 
This does not yield consistent improvements across models and benchmarks. Full table in Appendix.
}
\setlength{\tabcolsep}{0.35em}
\begin{tabular}{l|cc|cc|cc|cc}
\toprule
\textbf{Model} 
& \multicolumn{2}{c|}{\textbf{MMLU}} 
& \multicolumn{2}{c|}{\textbf{AI2-ARC}} 
& \multicolumn{2}{c|}{\textbf{GSM-Sym}} 
& \multicolumn{2}{c}{\textbf{SoEBench}} \\
\cmidrule(lr){2-3} \cmidrule(lr){4-5} \cmidrule(lr){6-7} \cmidrule(lr){8-9}
 & Img & \makecell{$\Delta$} 
 & Img & \makecell{$\Delta$} 
 & Img & \makecell{$\Delta$} 
 & Img & \makecell{$\Delta$} \\
\midrule
Deepseek-Small & 51.2 & {\color{red}-13.1} & 70.0 & {\color{red}-12.8} & 60.9 & {\color{red}-3.4} & 4.7 & {\color{red}-1.3} \\
GPT-4o-mini & 77.0 & {\color{red}-5.0} & 89.9 & {\color{red}-2.1} & 89.0 & {\color{red}-0.4} & 62.0 & {\color{mygreen}+2.7} \\
GPT-5-mini & 87.8 & {\color{mygreen}+1.2} & 93.3 & {\color{mygreen}+1.1} & 94.1 & {\color{mygreen}+0.8} & 97.3 & {\color{mygreen}+0.7} \\
Gemini-2.5 FL & 78.4 & {\color{red}-0.3} & 88.8 & {\color{mygreen}+0.2} & 50.5 & {\color{mygreen}+9.1} & 49.3 & {\color{red}-5.3} \\
Gemma-3-12B & 72.2 & {\color{red}-0.0} & 88.6 & {\color{mygreen}+1.0} & 88.1 & {\color{mygreen}+1.5} & 76.0 & {\color{red}-5.3} \\
Haiku-4.5  & 85.1 & {\color{mygreen}+0.1} & 92.7 & {\color{red}-0.7} & 94.1 & {\color{red}-1.3} & 97.3 & {\color{red}-1.3} \\
InternVL3-14B & 83.0 & {\color{red}-0.1} & 94.2 & 0.0 & 93.0 & {\color{red}-0.5} & 72.7 & {\color{red}-1.3} \\
Phi-4 & 31.4 & {\color{mygreen}+2.2} & 35.0 & {\color{mygreen}+9.0} & 54.5 & {\color{mygreen}+6.2} & 17.3 & {\color{red}-4.0} \\
Qwen-2.5-32B & 83.5 & {\color{red}-3.5} & 92.9 & {\color{red}-2.5} & 93.0 & {\color{mygreen}+0.4} & 75.2 & {\color{mygreen}+10.1} \\
\bottomrule
\end{tabular}
\label{tab:ocr_first}
\end{table}

\paragraph{OCR first, then solve}
\citet{chen2024omnixr} demonstrate improved results when instructing MLLMs to first extract information from an image before solving the task. Table \ref{tab:ocr_first} reports the accuracy difference between direct image processing and the OCR-first approach. Some models improve on many, if not all, \textbf{REST} benchmarks (e.g., GPT-5-mini, Phi-4 and Gemma-3-4B). 
However, the performance is also hurt substantially in some cases (e.g., InternVL3 2B on~\textsc{SoEBench}). 
These results reinforce that OCR is not the confounding factor behind cross-model inconsistency, because the OCR-first approach does not yield consistent results and can actually hurt performance.

\begin{table}[t]

\centering
\footnotesize 
\caption{\textbf{REST+ scores for 15 MLLMs show varying degrees of cross-modal inconsistency.} RER and CFR scores are given for all questions and for those where models obtain correctly recognised the rendered text (OCR$\checkmark$).}
\setlength{\tabcolsep}{0.45em}
\begin{tabular}{l|cc|ccc}
\toprule
\textbf{Model} 
& \multicolumn{2}{c|}{\makecell{\textbf{REST+} (OCR\checkmark)}} 
& \multicolumn{3}{c}{\makecell{\textbf{REST+} (All Questions)}} \\
\cmidrule(lr){2-3} \cmidrule(lr){4-6}
 & RER $
 \uparrow$ & CFR $
 \downarrow$ & RER $
 \uparrow$ & CFR $
 \downarrow$ & OCR $
 \uparrow$ \\
 \midrule
Deepseek-Tiny & 5.8 & 96.2 &  3.5 & 97.0 & 82.7 \\
Phi-4 & 11.3 & 89.2 &  7.6 & 90.5 & 80.1 \\
Phi-3.5 & 14.1 & 91.7 & 7.6 & 93.8 & 77.3 \\
Deepseek-Small & 24.5 & 76.9 & 21.0 & 78.9 & 91.5 \\
InternVL3 (2B) & 31.8 & 69.9 & 27.5 & 72.0 & 85.2 \\
Gemma-3 (4B) & 36.8 & 67.0 & 27.6 & 73.9 & 70.5 \\
Gemma-3 (12B) & 53.2 & 46.7 & 49.9 & 49.5 & 86.6 \\
Qwen-2.5 (7B) & 53.9 & 45.8 & 49.7 & 49.7 & 91.7 \\
GPT-4o-mini & 61.2 & 38.8 & 60.8 & 39.0 & 95.7 \\
Mistral-Small & 61.6 & 38.4 & 55.4 & 44.2 & 89.6 \\
Gemini-2.5 Flash Lite & 63.8 & 32.4 & 62.7 & 33.3 & \textbf{95.9} \\
Haiku-4.5 (Claude) & 65.7 & 34.0 & 45.8 & 54.0 & 82.6 \\
GPT-5-mini & 67.6 & 32.4 &  64.8 & 35.1 & 94.5 \\
Qwen-2.5 (32B) & 69.4 & 30.4 & 65.5 & 33.7 & 90.7 \\
InternVL3 (14B) & \textbf{72.1} & \textbf{27.9} & \textbf{69.3} & \textbf{30.2} & 88.8 \\
\bottomrule
\end{tabular}
\label{tab:rest_plus_consistency}
\end{table}

\section{Benchmarking Results: REST+}
\label{ch:results-plus}
We evaluate the same MLLMs on our more challenging \textbf{REST+} benchmark, which contains image permutations with different fonts, resolution and colours (see Table~\ref{tab:rest_plus_consistency}). First, OCR performance is worse than for REST, as expected, because of the lower image resolutions. Second, the overall RER and CFR scores are lower for \textbf{REST+} than the basic version. The high CFR values (best: 30.2\%) indicate that the input format matters: a different font, colour, or resolution can determine whether an answer is correct. Finally, the model rankings differ from the OCR correct and the full set of questions. In particular, InternVL3-14B achieves the highest consistency scores on both, but others (Claude Haiku and Mistral) decrease significantly when considering the full set of questions.

\begin{table}[t]
\centering
\footnotesize 
\caption{\textbf{Cross-modal consistency varies across font resolution.} RER and OCR across DPI levels show that some models maintain consistent performance, while others do not (set of all questions). 
}
\setlength{\tabcolsep}{0.45em}
\begin{tabular}{l|cc|cc|cc}
\toprule
\textbf{Model} 
& \multicolumn{2}{c|}{\textbf{DPI@50}} 
& \multicolumn{2}{c|}{\textbf{DPI@100}} 
& \multicolumn{2}{c}{\textbf{DPI@200}} \\
\cmidrule(lr){2-3} \cmidrule(lr){4-5} \cmidrule(lr){6-7}
& {RER} $\uparrow$ & {OCR} $\uparrow$
& {RER} & {OCR}  
& {RER} & {OCR} \\
\midrule
Haiku-4.5 (Claude) & 53.6 & 57.4 & 74.4 & 92.2 & \textbf{76.3} & \textbf{94.3} \\
Qwen-2.5 (7B) & 58.1 & 79.1 & 66.5 & 97.0 & \textbf{69.8} & \textbf{97.2} \\
Gemma-3 (12B) & 62.5 & 73.9 & \textbf{64.9} & 90.9 & 63.6 & \textbf{92.9} \\
Mistral-Small & 60.0 & 73.9 & 76.9 & 95.8 & \textbf{78.4} & \textbf{96.6} \\
GPT-4o-mini & 70.8 & 93.5 & \textbf{73.2} & 96.4 & 72.2 & \textbf{96.9} \\
Gemini-2.5 F Lite & 70.6 & 95.1 & 70.4 & 96.2 &  \textbf{70.9} & \textbf{96.3} \\
GPT-5-mini & 71.5 & 89.6 & 82.7 & 95.9 & \textbf{83.6} & \textbf{97.1} \\
Qwen-2.5 (32B) & 71.2 & 77.8 & 78.1 & 95.5 & \textbf{79.5} & \textbf{96.9} \\
InternVL3 (14B) & 75.3 & 88.9 & 79.0 & \textbf{89.1} & \textbf{79.9} & 88.6 \\
\bottomrule
\end{tabular}
\label{tab:ocr_rer_tab}
\vspace{-1em}
\end{table}

\subsection{RQ3a: Impact of resolution on inconsistency}
We investigate the effect of different resolutions on inconsistency (Table \ref{tab:ocr_rer_tab}). Some models perform consistently across different resolutions (InternVL3-14B and Gemini). However, others show decreased OCR and RER performance at DPI@50 (Claude and Gemma), indicating that OCR capabilities at this resolution are a key factor in inconsistency (if not controlling for OCR performance). Therefore, we conclude that OCR capabilities should be considered when evaluating cross-modal inconsistency (e.g., Claude Haiku would have shown worse performance if we had not controlled for incorrect OCR).

\begin{figure}[t]
    \centering
   \includegraphics[width=0.99\linewidth]{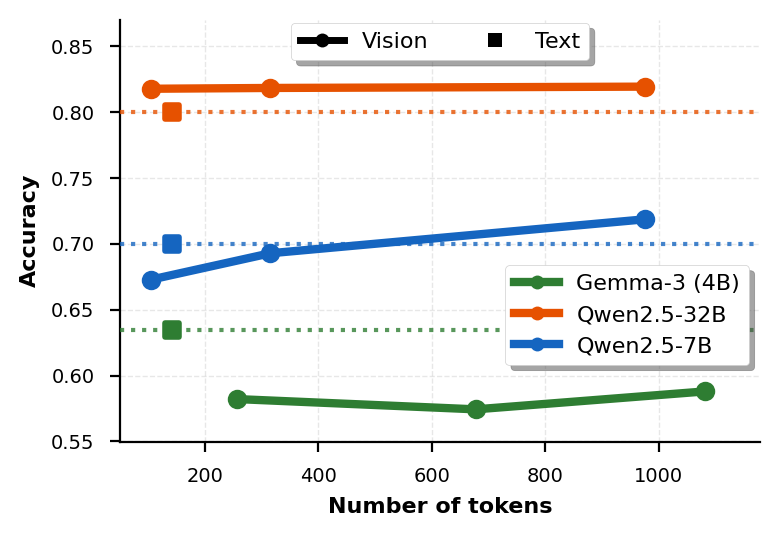}
    \caption{\textbf{Models generally achieve higher text accuracy despite using fewer text tokens.} Current MLLMs need more vision tokens than text tokens to achieve the same accuracy, except for Qwen2.5-VL-32B, where fewer vision tokens obtain higher accuracy (OCR-correct subset).}
    \label{fig:tokens_vs_accuracy}
    \vspace{-2em}
\end{figure}

\paragraph{Comparing efficiency of text and vision tokens} Different image resolutions result in different numbers of visual tokens for many models. \citet{wei2025deepseek} have developed DeepSeek-OCR to maintain performance when compressing text into fewer visual tokens. We investigate whether current MLLMs could already obtain similar efficiency gains (where we use efficiency to describe obtaining similar or better performance with fewer tokens). 

Figure \ref{fig:tokens_vs_accuracy}  shows text and image accuracy for three models at different DPI levels (50, 100, 200). Qwen2.5-32B demonstrates higher efficiency for vision, where less image tokens achieve 1.8\% higher accuracy than text tokens (104 image tokens on average ($\sigma$ = 24) at 50 DPI, vs. 142 text tokens). In contrast, Qwen2.5-7B shows less visual token efficiency at DPI@50, where text and vision accuracies differ significantly despite comparable token counts. In fact, only Qwen-32B exhibits the property of being more efficient with vision tokens while scoring better than with text tokens, all other models (see Appendix) either perform better with text or are more efficient with text. Note that image token counts exclude text instruction tokens.
Interestingly, InternVL3-14B (the most consistent model on \textbf{REST+}) uses approximately 1600 visual tokens (regardless of DPI), while requiring only 160 text tokens on average, a 10:1 ratio. Although InternVL3 shows higher image accuracy than text accuracy, text tokens are more efficient. Since we lack exact token usage data for closed-source models, we cannot perform the same analysis. Nevertheless, Claude, the GPTs, and Gemini all obtain better performance for text than image accuracy at DPI@50.

DeepSeek-OCR presents an interesting direction for future research in making MLLMs more efficient. However, we observe that current models reason substantially better through the text modality and that fewer text tokens than vision tokens are needed to obtain the same performance (except for Qwen2.5-VL-32B). 
In other words, although models can correctly read text from images, they do not necessarily reason as well as with conventional text input.

\begin{figure}[t]
    \centering
    \includegraphics[width=\linewidth]{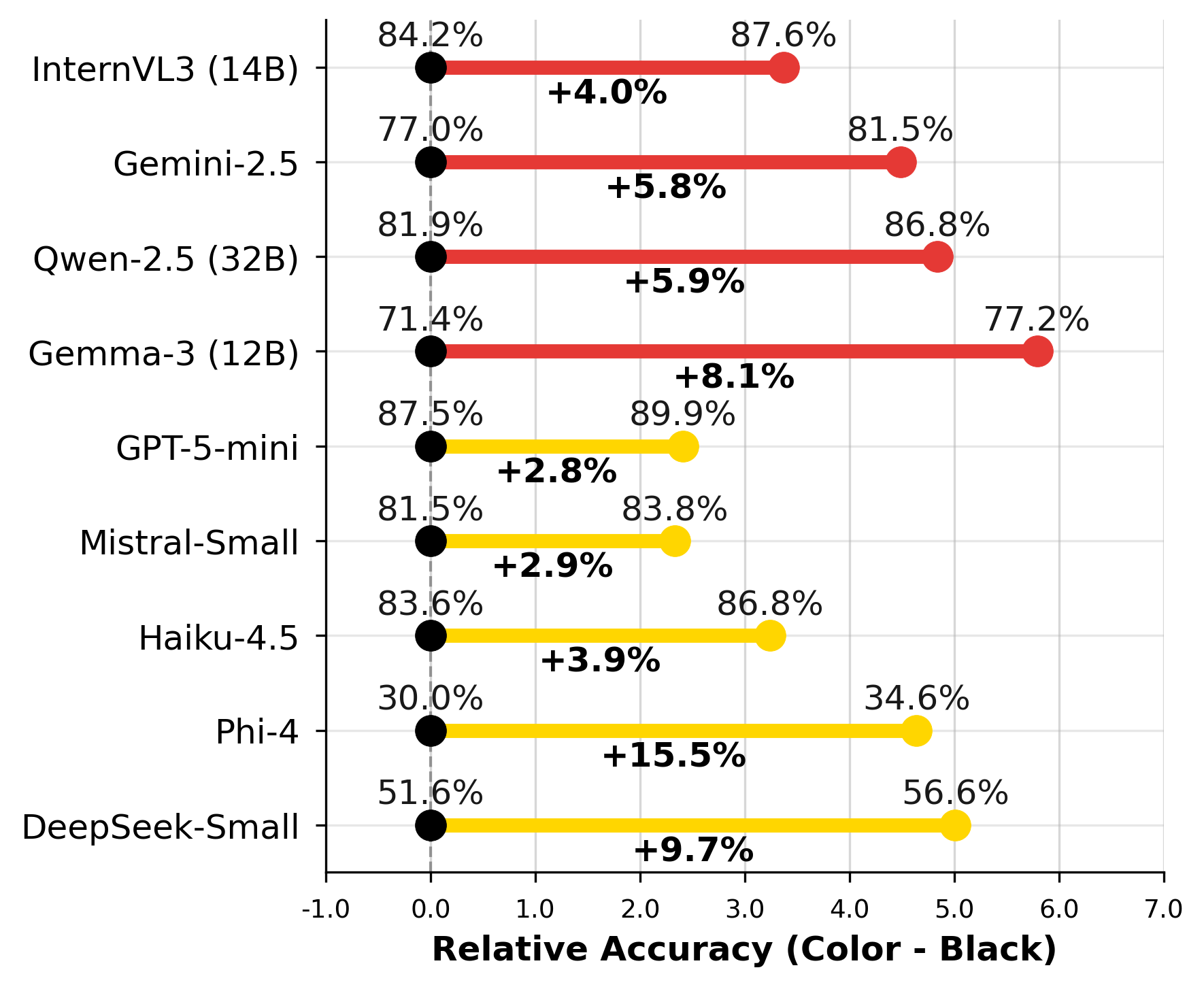}
    \caption{\textbf{Colored text makes models perform better.} Relative improvements from either red or yellow text compared to black (OCR-correct subset).}
    \label{fig:color_performance}
\end{figure}

\subsection{RQ3b: Impact of font, colour on inconsistency}
Surprisingly, font families show no clear differences in image accuracy, despite our initial expectations that cursive fonts would be harder to read. Most models stay within 2\% absolute difference between fonts (only Phi-3.5 shows a 5.3\% difference between DejaVu Sans and Courier New).
Unexpectedly, most models perform better with coloured text than black text (Figure~\ref{fig:color_performance}, DPI@200). Most MLLMs achieve higher accuracies with red or yellow fonts. Multiple models even obtain more than 5\% relative improvement (e.g. DeepSeek-Small and Qwen-32B). We show more detailed results in the Appendix.

\section{RQ4: Multimodal Representation Analysis}
To answer \textbf{RQ4}, we aim to find the internal mechanisms that determine a model's inconsistency. We postulate that models with higher similarities between their modalities score better on our benchmark. We calculate the \textit{implicit alignment score} introduced by Shukor and Cord \cite{shukor2024implicit}. 
They define alignment in terms of cosine similarity: the higher the similarity score, the more the representations point in the same direction. They find that this score is a proxy metric for task performance on several multimodal benchmarks. 

\begin{figure}[t!]
\centering
\vspace{-15pt}
\begin{subfigure}{0.3\columnwidth}
\centering
    \fbox{\includegraphics[width=0.5\textwidth]{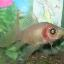}}
    \caption{Image}
\end{subfigure}
\begin{subfigure}{0.3\columnwidth}
\centering
    \fbox{\includegraphics[width=.8\textwidth]{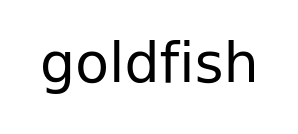}}
    \caption{Written-down}
\end{subfigure}
\begin{subfigure}{0.3\columnwidth}
    \parbox[c][1.5cm][c]{\linewidth}{\centering `goldfish'}
    \caption{Text}
\end{subfigure}
\caption{Samples for evaluating the representations of MLLMs.}
\label{fig:imagenet}
\end{figure}

We use Imagenet \cite{deng2009imagenet}, containing natural images and text labels. We expand the data by generating images containing written-down labels (see Figure \ref{fig:imagenet}). This simulates our \textbf{REST} benchmark which contains text-only, image-only and mixed data. We perform the experiments on 1,000 samples. Note that we exclude our \textbf{REST} benchmark, because our aim is to assess the correlation between our benchmark and the modality gap (\textbf{RQ4}). To this end, we analyse real-world images which are expected to be more familiar to MLLMs and yield in-distribution image representations. Note that natural images contain visual details such as depth and layout with no precise text counterpart. Therefore, we verify that inconsistency indeed extends beyond typographic inputs by additionally constructing a same-content setting using chess positions (App \ref{app:chess}).

\begin{figure*}[htb]
    \centering
    \includegraphics[width=.99\linewidth]{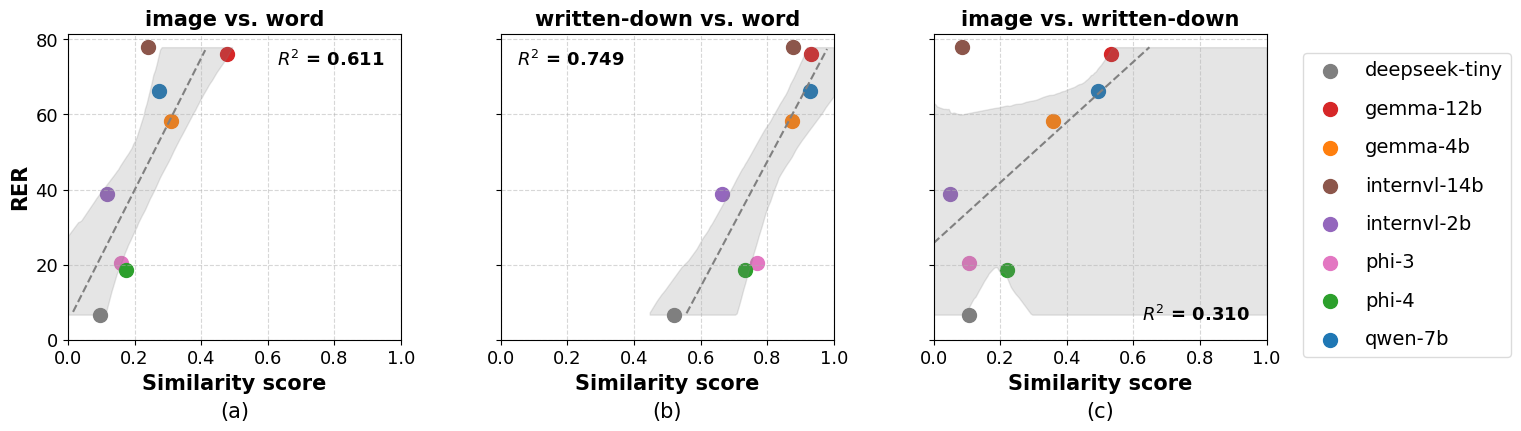}
    \caption{\textbf{Benchmark performance is correlated to the similarity of modalities.} The similarity between image vs. word (a) and written-down vs. word (b) representations correlates with RER (as determined using our \textbf{REST} benchmark). $R^2$ denotes the variance explained by the fitted line, and the grey area shows the bootstrapped 95\% confidence interval.}
    \label{fig:similarity}
    \vspace{-1em}
\end{figure*}
Since we need to retrieve the hidden activations, we focus on the open-source MLLMs from our previous experiments. 
We use three prompts, depending on the modality (1) images: “The main object in this image can be described as:”, (2) written-down images: “Repeat the text on the image:”, and (3) text: “Repeat the text \{text\_prompt\}:”.

These prompts are chosen such that the model is prompted to (1) give the same response in all modalities and (2) pay attention to the image or text of interest. For our analysis, we represent each sequence of tokens by computing its mean embedding. Given image tokens $\mathbf{I} = \{\mathbf{i_1}, \mathbf{i_2}, ..., \mathbf{i_A}\} \in \mathbb{R}^{A \times d}$ where $A$ denotes the number of image tokens and $d$ is the embedding dimension, we compute their mean representation $\mathbf{\overline{i}} = \frac{1}{A}\sum_{k=1}^A \mathbf{i}_k \in \mathbb{R}^d$. Similarly, for text tokens $\mathbf{T} = \{\mathbf{t_1}, \mathbf{t_2}, ..., \mathbf{t_B}\} \in \mathbb{R}^{B \times d}$ where $B$ denotes the number of text tokens, we compute $\mathbf{\overline{t}} = \frac{1}{B}\sum_{k=1}^B \mathbf{t}_k \in \mathbb{R}^d$. 
We measure the similarity between image and text representations using cosine similarity:
\begin{equation}
\operatorname{sim}(\mathbf{I}, \mathbf{T}) = \frac{\mathbf{\overline{i}} \cdot \mathbf{\overline{t}}}{\|\mathbf{\overline{i}}\| \|\mathbf{\overline{t}}\|}.
\end{equation}

To evaluate cross-modal alignment, we compute a retrieval accuracy score that measures how often each sample correctly retrieves its corresponding pair from the opposite modality. For a dataset containing $N$ paired image-text samples, we determine for each image $i$ which text $j$ has the highest similarity:
\begin{equation}
y_i = \operatorname{argmax}_{j \in \{1,...,N\}} \operatorname{sim}(\mathbf{I}_i, \mathbf{T}_j).
\end{equation}

The similarity score is then computed as the fraction of correct retrievals:
\begin{equation}
\text{Similarity score} = \frac{1}{N} \sum_{i=1}^N \mathbbm{1}[y_i = i],
\end{equation}
where $\mathbbm{1}[\cdot]$ is the indicator function that equals 1 when the retrieved text index matches the correct pairing and 0 otherwise. This metric effectively measures the accuracy of cross-modal retrieval, where higher scores indicate better alignment between the image and text representations. We compute this similarity score at each layer of each model and report the maximum score across all layers. We repeat this process bidirectionally (image-to-text and text-to-image retrieval), yielding multiple similarity scores per model that characterize the cross-modal alignment from different perspectives.
Figure~\ref{fig:similarity} shows that similarity scores are correlated with the Render-Equivalence Rate (RER), especially for the similarity between images and words and between written-down labels and words. This indicates that matching questions are very similar to each other in consistent models (i.e. models with high RER). Note that the similarity scores between written-down labels and words are higher overall, indicating that written-down representations are more similar to matching word representations than image representations are.

\section{Related Work}
Most MLLM benchmarks are visual question-answering tasks~\cite{goyal2017vqav2, hudson2019gqa, singh2019textvqa, yue2024mmmu, liu2023mmbench, lu2023mathvista} and do not include data samples where the same question is presented across different modalities. The Omni-R benchmark~\citep{chen2024omnixr} does contain paired questions in text, image, video, and audio, based on MMLU-Pro~\citep{wang2024mmlu}. However, it evaluates inconsistency only in closed-source MLLMs and does not control for OCR effects.
MMMU-Pro~\cite{yue2024mmmupro} contains a vision-only subtask in which the text is rendered alongside the original visuals. Our work differs in making the OCR task as simple as possible, whereas MMMU-Pro renders text in screenshots or natural photos. The vision-language consistency dataset of~\citet{zhang2024crossmodalconsistencymultimodallarge} includes multiple existing text benchmarks, but evaluates only GPT-4V~\cite{achiam2023gpt} and similarly does not control for OCR. MMIR~\citep{yan2025multimodal} focuses on a different form of multimodal inconsistency: detecting semantic mismatches between text and vision. \citet{alonso2025vision} show that MLLMs are unable to consistently match the same entity across modalities, which contain complementary information, and not identical (like ours). Similarly, \citet{samson2024privacy} find cases where MLLMs handle the concept of privacy differently in text and vision, and \citet{sim2025can} study modality collapse (ignoring certain modalities) and find that the text modality often dominates.

Current literature agrees on the existence of a modality gap in different MLLMs, where text and image embeddings occupy different regions in the joint embedding space \cite{liang2022mind, shukor2024implicit, papadimitriou2025interpreting, shu2025large}. Liang et al. show the modality gap for the first time for CLIP, which they attribute to the two modalities initialising a narrow embedding cone, which is preserved by the loss function \citep{liang2022mind}. This phenomenon has since been documented more broadly across MLLMs \citep{shukor2024implicit}. Shukor and Cord actually find a positive correlation between the cosine similarity of the text and image representations within one prompt and task performance \cite{shukor2024implicit}. Eslami and de Melo find that mitigating this gap in CLIP (by parameter sharing) improves the performance on downstream tasks \cite{eslami2024mitigategapinvestigatingapproaches}.

\section{Discussion and Conclusion}

In this work, we present the \textbf{REST} and \textbf{REST+} benchmarks to evaluate cross-modal inconsistency in MLLMs. 
Through our experiments, we find that there is considerable cross-modal inconsistency in state-of-the-art MLLMs, which answers \textbf{RQ1}. 
This varies across models, with cross-modal inconsistency scores ranging from 10\% to 90\%, depending on the MLLM. 
We find that GPT-5-mini and Claude Haiku 4.5 achieve the highest consistency scores and that almost all MLLMs perform best in the text modality. 

Despite only evaluating on questions with correct OCR, we find that the cross-modal inconsistency persists. 
This indicates that cross-modal inconsistency is not simply a by-product of poor OCR performance, which answers \textbf{RQ2}. 
We find that neither rendering images as text (OCR first) nor rendering text as images solves the inconsistency. 
We find that font resolution has an effect on cross-modal inconsistency, where higher DPI correlates with more consistent MLLM performance across modalities, even when controlling for OCR performance. This answers \textbf{RQ3a}. 
In other words, although models can correctly read text from images, they do not necessarily reason as well as they would with conventional text input.
We show that font does not have a clear impact on cross-modal inconsistency, while text colour does (\textbf{RQ3b}). 
In our multimodal representation analysis experiment, we examine the internal representations of the same concept in different modalities. 
We observe a positive correlation between the cross-modal similarity of internal representations and our RER consistency score. 
This implies that the direction of representations from matching samples is linked to modality inconsistency, which answers \textbf{RQ4}. 
Future work could study whether this effect is causal, in particular by optimising the representations to be more similar and observing whether the RER consistency score increases.

{
    \small
    \bibliographystyle{ieeenat_fullname}
    \bibliography{main}
}

\clearpage
\setcounter{page}{1}
\maketitlesupplementary

\section{Benchmark Implementation Details}
We provide complete specifications for reproducing the \textbf{REST} benchmark experiments, including prompt templates and dataset examples. All code, data, and model outputs will be released publicly upon acceptance to facilitate future research on cross-modal consistency.

\subsection{REST}
\begin{figure*}[htb]
    \centering

    \begin{subfigure}[b]{0.49\textwidth}
        \centering
        \frame{\includegraphics[width=\textwidth]{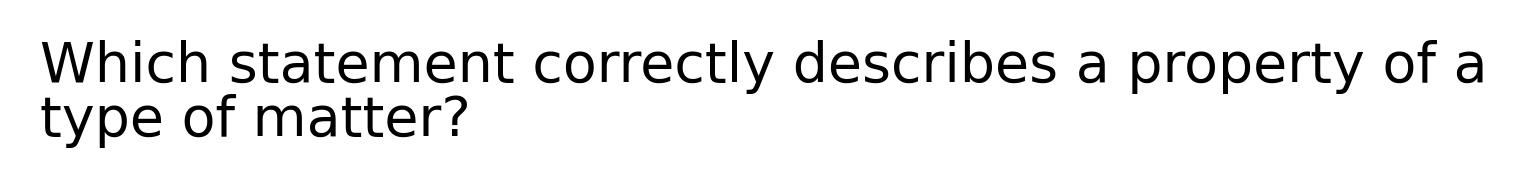}}
        \caption{ARC - Mixed modality}
        \label{fig:arc_mixed}
    \end{subfigure}
    \hfill
    \begin{subfigure}[b]{0.49\textwidth}
        \centering
        \frame{\includegraphics[width=\textwidth]{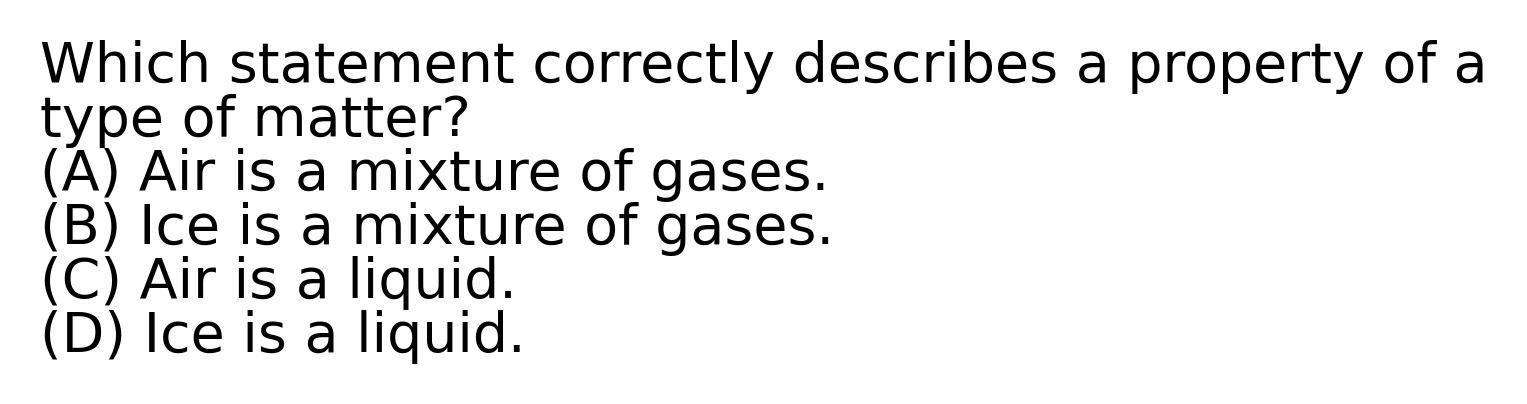}}
        \caption{ARC — Image modality}
        \label{fig:arc_image}
    \end{subfigure}

    \vspace{1em}

    \begin{subfigure}[b]{0.49\textwidth}
        \centering
        \frame{\includegraphics[width=\textwidth]{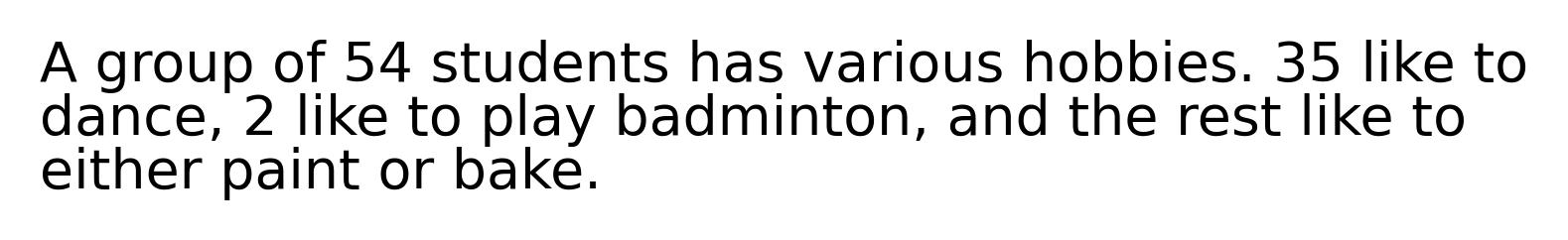}}
        \caption{GSM8K — Mixed modality}
        \label{fig:gsm_mixed}
    \end{subfigure}
    \hfill
    \begin{subfigure}[b]{0.49\textwidth}
        \centering
        \frame{\includegraphics[width=\textwidth]{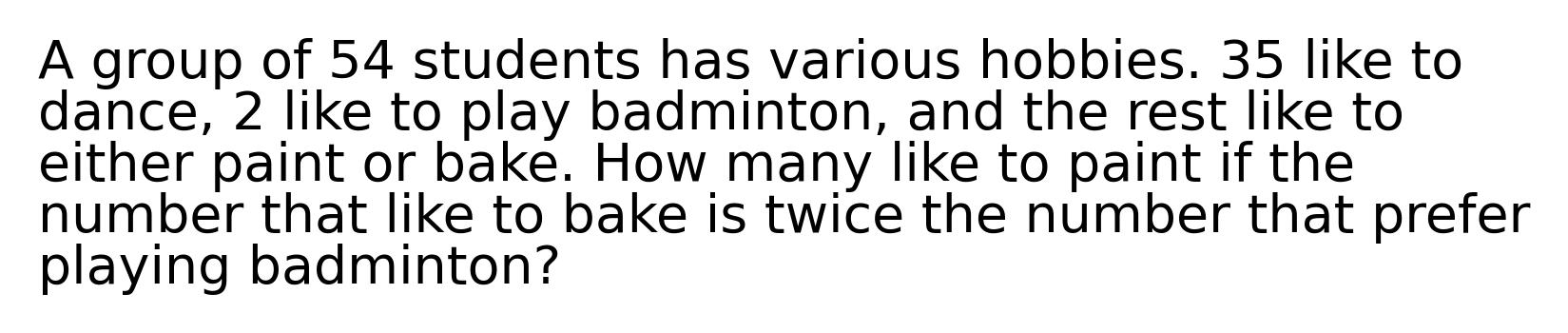}}
        \caption{GSM8K — Image modality}
        \label{fig:gsm_image}
    \end{subfigure}

    \vspace{1em}

    \begin{subfigure}[b]{0.49\textwidth}
        \centering
        \frame{\includegraphics[width=\textwidth]{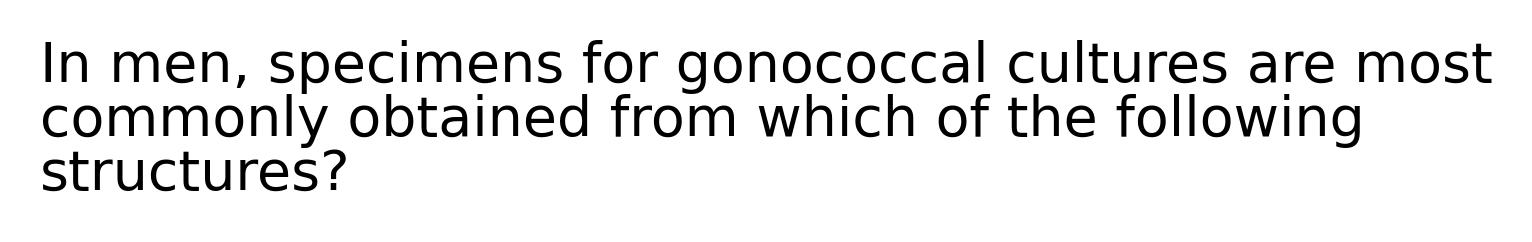}}
        \caption{MMLU — Mixed modality}
        \label{fig:mmlu_mixed}
    \end{subfigure}
    \hfill
    \begin{subfigure}[b]{0.49\textwidth}
        \centering
        \frame{\includegraphics[width=\textwidth]{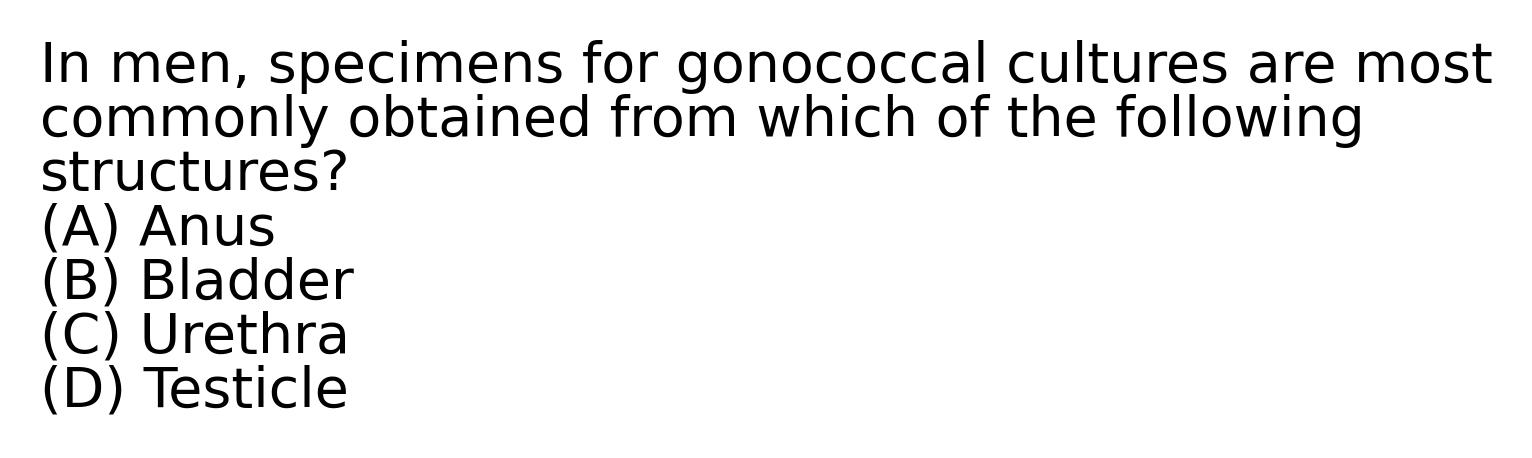}}
        \caption{MMLU — Image modality}
        \label{fig:mmlu_image}
    \end{subfigure}

    \caption{Examples of ARC, GSM8K, and MMLU questions in mixed and image modalities from our REST benchmark. In the mixed modality, part of the content (e.g., multiple-choice options or context) is provided as text while the rest is rendered as an image. In the image modality, the entire content is rendered as a single image.}
    \label{fig:combined_modality_examples}
\end{figure*}

\paragraph{Images}
In Figure \ref{fig:combined_modality_examples}, we show examples of both mixed and image modalities for randomly selected questions from MMLU, ARC, and GSM8K-Symbolic. Samples for \textsc{SoEBench} are provided in the following section. We render all images at DPI 200 on a white background using black DejaVu Sans font to maximise readability and ensure consistent OCR performance. To minimise OCR complexity, we exclude questions exceeding 800 characters and those containing LaTeX formatting, as mathematical notation can introduce ambiguity in text recognition tasks. The mixed modality format depends on the benchmark structure: for multiple-choice questions (MMLU, ARC), the answer options appear as text while the question context is rendered as an image; for open-ended tasks (GSM8K-Symbolic, \textsc{SoEBench}), the problem context appears as an image while the final question is presented as text.

\paragraph{Prompts} 
Figure \ref{fig:prompt_templates} presents the prompt templates for the four core tasks in the \textbf{REST} benchmark. We encourage Chain-of-Thought reasoning across all modalities and maintain consistent instructions, only adjusting for how a model should solve a question. For the OCR task, we explicitly instruct models to perform only text transcription without solving the problem, as some models would otherwise attempt to provide answers alongside the transcription.

\begin{figure*}[htb]
\centering
\begin{subfigure}[b]{0.48\textwidth}
\centering
\begin{tcolorbox}[
    colback=cvprblue!90,
    coltext=white,
    colframe=white!20,
    boxrule=0.4pt,
    arc=2pt,
    left=6pt,
    right=6pt,
    top=6pt,
    bottom=8pt,
    enhanced jigsaw,
    width=\textwidth,
]
\small
Solve the following question.\\[0.5em]
Think step by step, but put the answer (A, B, C or D) on the very last line,
preceded by \texttt{Answer:}.  
Do not write anything else on that line.\\[0.9em]
\textbf{Example:}\\[-0.2em]
Reasoning\dots\\
Answer: A\\\\
\colorbox{white}{%
  \parbox{\dimexpr\linewidth-2\fboxsep}{
      \small \textcolor{black}{\texttt{Question}}
  }%
}
\end{tcolorbox}
\caption{Text modality}
\end{subfigure}
\hfill
\begin{subfigure}[b]{0.48\textwidth}
\centering
\begin{tcolorbox}[
    colback=cvprblue!90,
    coltext=white,
    colframe=white!20,
    boxrule=0.4pt,
    arc=2pt,
    left=6pt,
    right=6pt,
    top=6pt,
    bottom=8pt,
    enhanced jigsaw,
    width=\textwidth,
]
\small
Solve the question in the image.\\[0.5em]
Think step by step, but put the answer (A, B, C or D) on the very last line,
preceded by \texttt{Answer:}.  
Do not write anything else on that line.\\[0.9em]
\textbf{Example:}\\[-0.2em]
Reasoning\dots\\
Answer: C
\end{tcolorbox}
\caption{Image modality}
\end{subfigure}

\vspace{0.5em}

\begin{subfigure}[b]{0.48\textwidth}
\centering
\begin{tcolorbox}[
    colback=cvprblue!90,
    coltext=white,
    colframe=white!20,
    boxrule=0.4pt,
    arc=2pt,
    left=6pt,
    right=6pt,
    top=6pt,
    bottom=8pt,
    enhanced jigsaw,
    width=\textwidth,
]
\small
Read the question in the image and choose from the options below.\\[0.5em]
Think step by step, but put the answer (A, B, C or D) on the very last line,
preceded by \texttt{Answer:}.  
Do not write anything else on that line.\\[0.9em]
\textbf{Example:}\\[-0.2em]
Reasoning\dots\\
Answer: B \\\\
\colorbox{white}{%
  \parbox{\dimexpr\linewidth-2\fboxsep}{
      \small \textcolor{black}{\texttt{Multiple Choice Options}}
  }%
}
\end{tcolorbox}
\end{subfigure}
\hfill
\begin{subfigure}[b]{0.48\textwidth}
\centering
\begin{tcolorbox}[
    colback=cvprblue!90,
    coltext=white,
    colframe=white!20,
    boxrule=0.4pt,
    arc=2pt,
    left=6pt,
    right=6pt,
    top=6pt,
    bottom=8pt,
    enhanced jigsaw,
    width=\textwidth,
]
\small
You are given an image that contains text. You must do the following:\\
1. Do not solve the question; just transcribe the text exactly as it appears.\\
2. Do not add extra commentary, only transcribe.\\
Please transcribe now.
\end{tcolorbox}
\caption{OCR verification}
\end{subfigure}
\caption{Prompt templates used in the \textbf{REST} benchmark for evaluating cross-modal consistency across MMLU, ARC, GSM-Symbolic. Each modality (text, image, mixed) receives task-specific instructions while maintaining consistent Chain-of-Thought reasoning requirements and standardized answer formatting. The OCR verification prompt (d) ensures that text recognition capabilities are assessed independently from reasoning performance. For non-MMLU benchmarks, prompts follow identical structures with minor adaptations: mixed modality varies based on whether questions are multiple-choice (options as text) or open-ended (questions as text, context as image).}
\label{fig:prompt_templates}
\end{figure*}

\subsection{REST+}
\begin{figure*}[t]
\centering
\begin{subfigure}[b]{0.14\textwidth}
    \centering
    \frame{\includegraphics[width=0.95\textwidth]{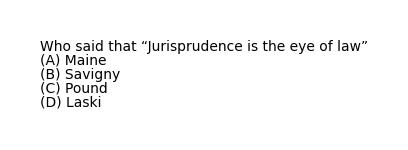}}
    \caption*{DejaVu Sans @ 50 DPI}
\end{subfigure}
\hfill
\begin{subfigure}[b]{0.28\textwidth}
    \centering
    \frame{\includegraphics[width=0.95\textwidth]{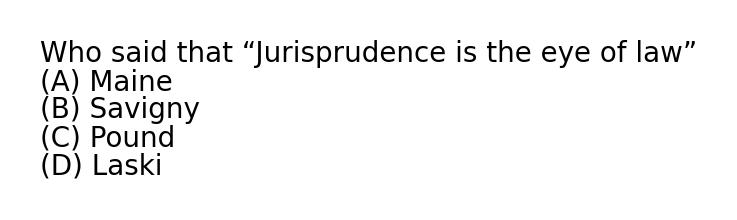}}
    \caption*{DejaVu Sans @ 100 DPI}
\end{subfigure}
\hfill
\begin{subfigure}[b]{0.57\textwidth}
    \centering
    \frame{\includegraphics[width=\textwidth]{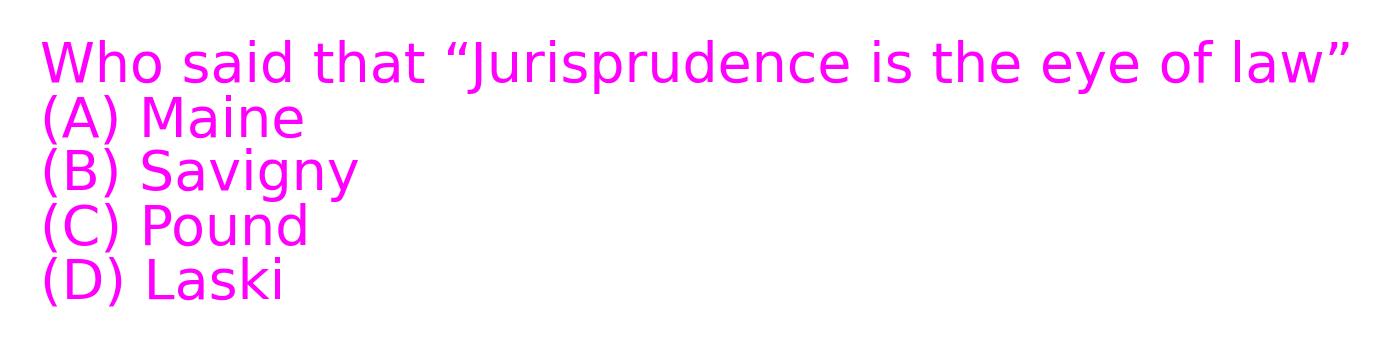}}
    \caption*{DejaVu Sans @ 200 DPI (Magenta \& Black)}
\end{subfigure}

\vspace{0.3cm}

\begin{subfigure}[b]{0.14\textwidth}
    \centering
    \frame{\includegraphics[width=0.95\textwidth]{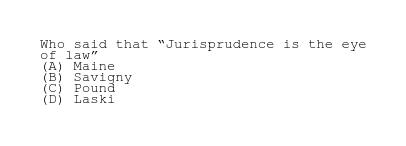}}
    \caption*{Courier New @ 50 DPI}
\end{subfigure}
\hfill
\begin{subfigure}[b]{0.28\textwidth}
    \centering
    \frame{\includegraphics[width=0.95\textwidth]{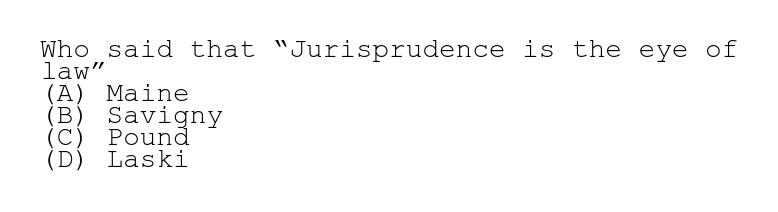}}
    \caption*{Courier New @ 100 DPI}
\end{subfigure}
\hfill
\begin{subfigure}[b]{0.57\textwidth}
    \centering
    \frame{\includegraphics[width=\textwidth]{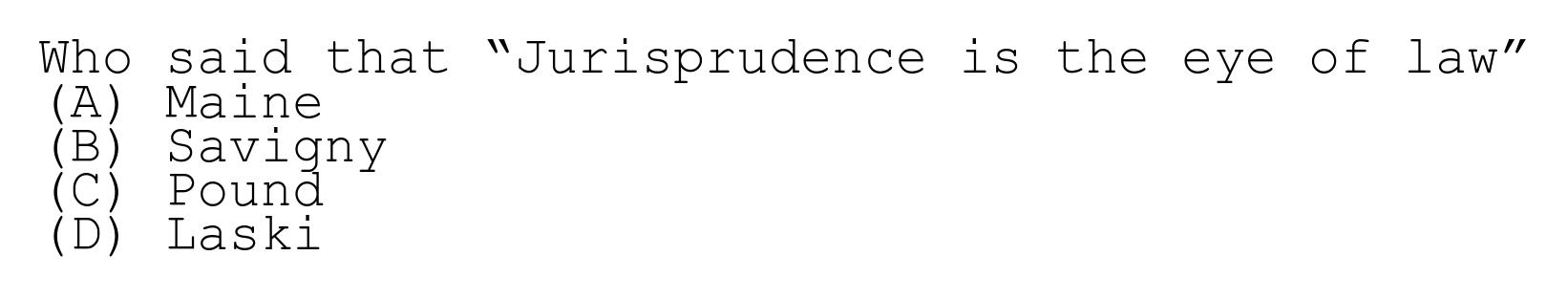}}
    \caption*{Courier New @ 200 DPI}
\end{subfigure}

\vspace{0.3cm}

\begin{subfigure}[b]{0.14\textwidth}
    \centering
    \frame{\includegraphics[width=0.95\textwidth]{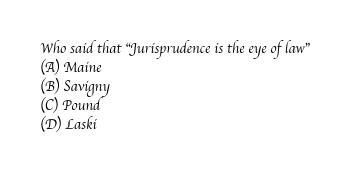}}
    \caption*{Cursive @ 50 DPI}
\end{subfigure}
\hfill
\begin{subfigure}[b]{0.28\textwidth}
    \centering
    \frame{\includegraphics[width=0.95\textwidth]{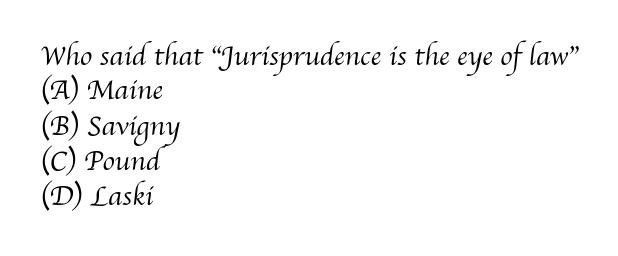}}
    \caption*{Cursive @ 100 DPI}
\end{subfigure}
\hfill
\begin{subfigure}[b]{0.57\textwidth}
    \centering
    \frame{\includegraphics[width=\textwidth]{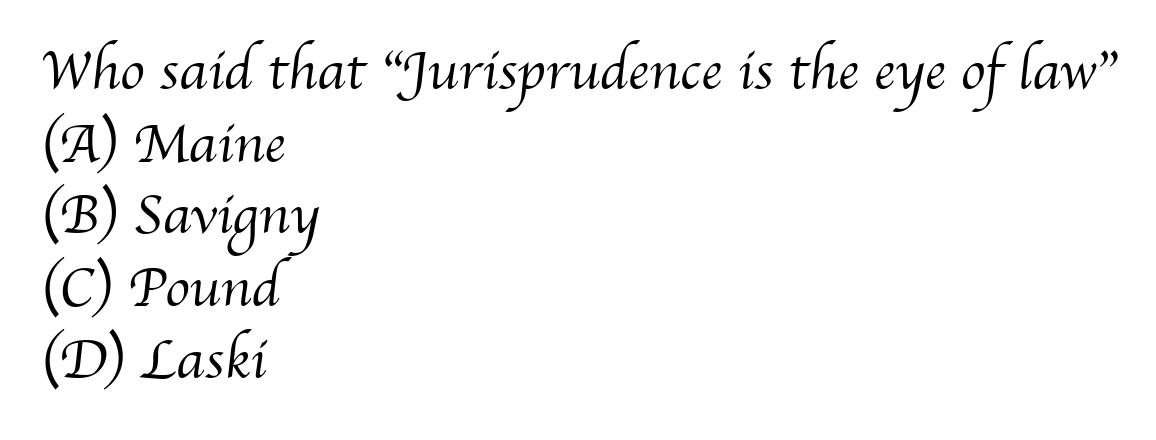}}
    \caption*{Cursive @ 200 DPI (Magenta)}
\end{subfigure}

\caption{Visual permutations in REST+ benchmark showing the same MMLU question rendered with varying resolutions (columns: 50, 100, 200 DPI) and fonts (rows: DejaVu Sans, Courier New, Cursive). We manually verified that images are legible at DPI 50.}
\label{fig:rest_plus_grid}
\end{figure*}

For the more challenging benchmark, we create 10 image permutations per question (Figure \ref{fig:rest_plus_grid}). We generate 9 combinations using three font families (DejaVu Sans, Courier New, and Cursive) and three resolutions (50, 100, and 200 DPI). We specifically chose these distinct font types to evaluate how cross-modal consistency varies across different typographic styles. Additionally, we create one color variant using DejaVu Sans at 200 DPI, with colors assigned cyclically using modulo operation from a set of six standard colors (red, green, blue, cyan, magenta, and yellow). We manually verified that text remains legible at 50 DPI for all font families to ensure that performance differences reflect reasoning capabilities rather than readability issues.

We limit our evaluation to these 10 permutations for computational feasibility, as each question requires 21 forward passes (1 text, 10 OCR verification, and 10 image-based solving). Questions are sampled from MMLU at 10\% per subject area to maintain subject balance across the reduced dataset. With 1,085 questions and 21 forward passes per question, we perform 22,785 evaluations per MLLM. We exclude the mixed modality from REST+ as it would require 10 additional forward passes per question, substantially increasing computational costs.

For the prompts, we use the same templates as in \textbf{REST} depicted in Figure \ref{fig:prompt_templates}, maintaining consistency in evaluation instructions across both benchmarks.

\subsection{\textsc{SoEBench}}
To ensure zero data contamination and minimise OCR complexity, we introduced \textsc{SoEBench}, a novel system-of-equations benchmark specifically designed for evaluating cross-modal consistency. Each puzzle presents $n$ letter variables (A through E for $n \in \{3, 4, 5\}$) that must be solved to find integer values between 1 and 9. The benchmark consists of 150 puzzles total, with 50 puzzles for each value of $n$.

Each puzzle contains $n$ clue equations plus one final equation to solve. Variables appear with coefficients ranging from 1 to 3, combined using basic arithmetic operations (+, -, *). For example, a clue equation might read ``2A + B = 15" while the final equation presents ``3B - A + 2C = ?" where the model must determine the result. We ensure each puzzle has a unique solution by verifying that only one assignment of values satisfies all clue equations simultaneously.

For the mixed modality, we render the clue equations as an image while presenting the final equation as text. For the image modality, all equations, including the final question, are rendered together. We use DejaVu Sans font at 200 DPI on a white background with black text to maximise readability. The restricted symbol set (digits 0-9, letters A-E, and basic operators) ensures that OCR performance remains near-perfect for most models, allowing us to isolate reasoning capabilities from recognition challenges.

\begin{figure*}[htb]
\centering
\begin{subfigure}[b]{0.31\textwidth}
    \centering
    \frame{\includegraphics[width=\textwidth]{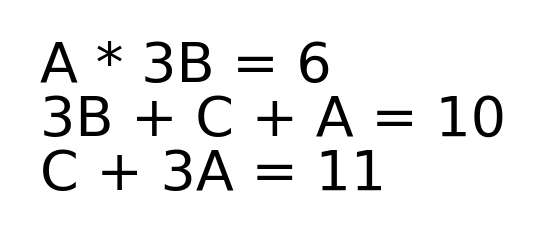}}
    \caption{3 Variables - Mixed modality}
\end{subfigure}
\hfill
\begin{subfigure}[b]{0.31\textwidth}
    \centering
    \frame{\includegraphics[width=\textwidth]{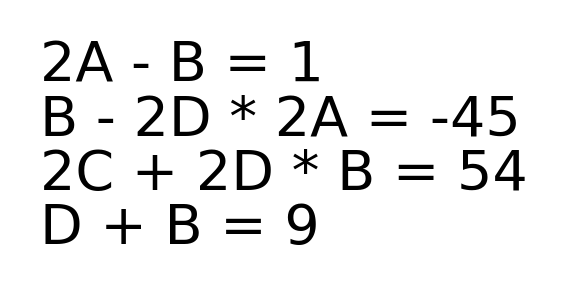}}
    \caption{4 Variables - Mixed modality}
\end{subfigure}
\hfill
\begin{subfigure}[b]{0.31\textwidth}
    \centering
    \frame{\includegraphics[width=\textwidth]{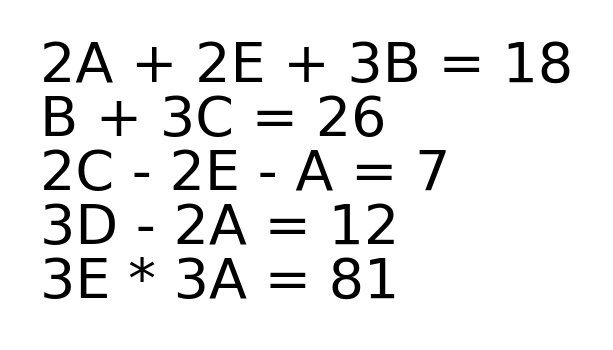}}
    \caption{5 Variables - Mixed modality}
\end{subfigure}

\begin{subfigure}[b]{0.31\textwidth}
    \centering
    \frame{\includegraphics[width=\textwidth]{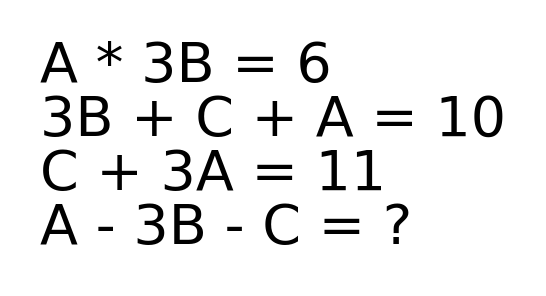}}
    \caption{3 Variables - Image modality}
\end{subfigure}
\hfill
\begin{subfigure}[b]{0.31\textwidth}
    \centering
    \frame{\includegraphics[width=\textwidth]{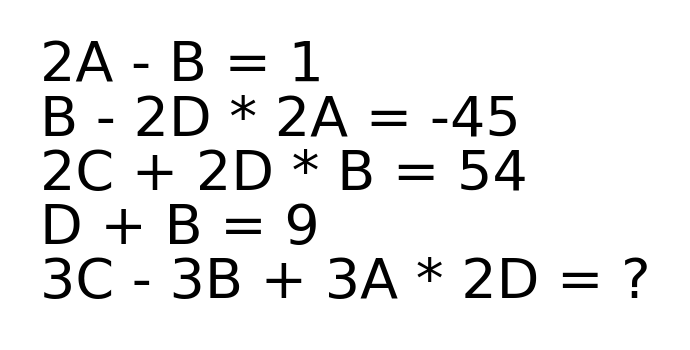}}
    \caption{4 Variables - Image modality}
\end{subfigure}
\hfill
\begin{subfigure}[b]{0.31\textwidth}
    \centering
    \frame{\includegraphics[width=\textwidth]{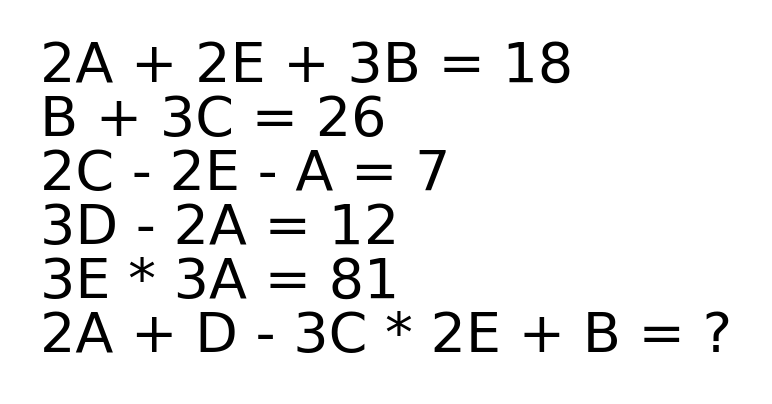}}
    \caption{5 Variables - Image modality}
\end{subfigure}

\caption{\textsc{SoEBench} examples with increasing complexity from 3 to 5 variables. Top row: mixed images containing only clue equations, used for the mixed modality where the final equation is presented as text. Bottom row: images including all equations with the final equation used for the image modality. Each puzzle requires finding integer values (1-9) for letter variables that satisfy all equations simultaneously. \textbf{Font rendering is consistent across all images (DejaVu Sans, 200 DPI); perceived size differences are due to automatic figure scaling for layout consistency.}}
\label{fig:soebench_examples}
\end{figure*}

Figure \ref{fig:soebench_examples} shows example puzzles with varying numbers of variables.

\paragraph{Prompts}
Figure \ref{fig:soebench_prompt_templates} presents the prompt templates. While the text, image, and mixed prompts follow similar Chain-of-Thought structures as \textbf{REST}, we introduce a delimiter-based output format using to clearly separate reasoning from the final answer, as models extensively reason. 

For the OCR verification task, we provide explicit formatting instructions requiring models to number each equation sequentially (1), (2), (3), etc. This structured format serves two purposes: it allows us to verify that models correctly identify the semantic structure of the equations, and it simplifies parsing and validation of OCR outputs, as we noticed different (correct) output formats for models.

\begin{figure*}[htb]
\centering
\begin{subfigure}[b]{0.48\textwidth}
\centering
\begin{tcolorbox}[
    colback=cvprblue!90,
    coltext=white,
    colframe=white!20,
    boxrule=0.4pt,
    arc=2pt,
    left=6pt,
    right=6pt,
    top=6pt,
    bottom=8pt,
    enhanced jigsaw,
    width=\textwidth,
]
\small
This puzzle contains equations with letters representing natural numbers from 1 to 9:\\[0.3em]
\colorbox{white}{%
  \parbox{\dimexpr\linewidth-2\fboxsep}{
      \small \textcolor{black}{\texttt{[All equations]}}
  }%
}\\[0.3em]
Provide the outcome of the final equation.\\[0.5em]
Think step by step. When completely finished, output:\\
\texttt{\#\#\#\#}\\
Then write only the final answer:\\
Example:\\
Reasoning step 1 $\cdots$ \\
Reasoning step 2 $\cdots$ \\
\texttt{\#\#\#\#}\\
\texttt{Answer: 28}
\end{tcolorbox}
\caption{Text modality}
\end{subfigure}
\hfill
\begin{subfigure}[b]{0.48\textwidth}
\centering
\begin{tcolorbox}[
    colback=cvprblue!90,
    coltext=white,
    colframe=white!20,
    boxrule=0.4pt,
    arc=2pt,
    left=6pt,
    right=6pt,
    top=6pt,
    bottom=8pt,
    enhanced jigsaw,
    width=\textwidth,
]
\small
This puzzle contains equations with letters representing natural numbers from 1 to 9.\\[0.3em]
Solve the puzzle in the image, provide the outcome of the final equation.\\[0.5em]
Think step by step. When completely finished, output:\\
\texttt{\#\#\#\#}\\
Then write only the final answer:\\
Example:\\
Reasoning step 1 $\cdots$ \\
Reasoning step 2 $\cdots$ \\
\texttt{\#\#\#\#}\\
\texttt{Answer: 28}
\end{tcolorbox}
\caption{Image modality}
\end{subfigure}

\vspace{0.5em}

\begin{subfigure}[b]{0.48\textwidth}
\centering
\begin{tcolorbox}[
    colback=cvprblue!90,
    coltext=white,
    colframe=white!20,
    boxrule=0.4pt,
    arc=2pt,
    left=6pt,
    right=6pt,
    top=6pt,
    bottom=8pt,
    enhanced jigsaw,
    width=\textwidth,
]
\small
This puzzle contains equations with letters representing natural numbers from 1 to 9:\\[0.3em]
Read the information in the image and then solve the equation below:\\[0.3em]
\colorbox{white}{%
  \parbox{\dimexpr\linewidth-2\fboxsep}{
      \small \textcolor{black}{\texttt{[Final equation with ?]}}
  }%
}\\[0.3em]
Think step by step. When completely finished, output:\\
\texttt{\#\#\#\#}\\
Then write only the final answer:\\
Example:\\
Reasoning step 1 $\cdots$ \\
Reasoning step 2 $\cdots$ \\
\texttt{\#\#\#\#}\\
\texttt{Answer: 28}
\end{tcolorbox}
\caption{Mixed modality}
\end{subfigure}
\hfill
\begin{subfigure}[b]{0.48\textwidth}
\centering
\begin{tcolorbox}[
    colback=cvprblue!90,
    coltext=white,
    colframe=white!20,
    boxrule=0.4pt,
    arc=2pt,
    left=6pt,
    right=6pt,
    top=6pt,
    bottom=8pt,
    enhanced jigsaw,
    width=\textwidth,
]
\small
You are given an image that contains a list of separate mathematical equations.\\[0.3em]
1. Do not solve or simplify these equations; just transcribe them exactly as they appear.\\
2. Retain the same order and use the following numbering, (1), (2), (3) per equation.\\
3. List 1 equation per item, for example (1) 3a + 2b + c = 11\\
4. Put each equation on its own line. \\
5. Use plain text as output, the operations that you can use are '*', '+', '-' and '='.\\
6. Do not add extra commentary-only transcribe equations.\\
Format your output like so:\\
(1) 2a + 3b = 10\\
(2) a + 3c = 30\\
(3) 2b + 5c = ?\\
Please transcribe now.
\end{tcolorbox}
\caption{OCR verification}
\end{subfigure}
\caption{Prompt templates used for \textsc{SoEBench} evaluation. Each modality receives specific instructions for solving systems of equations with letter variables. For the mixed modality, clue equations are provided as images while the final equation appears as text. For OCR, we instruct the models for a specific format, as models generate different types of correct output formats.}
\label{fig:soebench_prompt_templates}
\end{figure*}

\subsection{Model Configuration and Hardware}
We conduct all experiments using vLLM~\cite{kwon2023efficient} for computational efficiency. Temperature is set to 0 for deterministic outputs across all models where configurable. For proprietary models, we apply the following settings: GPT-5-mini uses minimal reasoning effort (temperature control unavailable), Claude Haiku-4.5 has thinking mode disabled, and Gemini-2.5 Flash Lite operates with thinking budget set to 0. Despite these computational optimisations, all models receive identical Chain-of-Thought prompting instructions to ensure fair comparison.

Open-source models follow vLLM's recommended configurations for optimal performance. Experiments run on single-GPU systems: NVIDIA RTX 6000 Ada (48GB VRAM) for most models, and NVIDIA H100 (80GB VRAM) for larger models (Mistral-Small-3.1-24B and Qwen2.5-32B) due to memory requirements.

\clearpage
\section{Extended Results} \label{app:extended}
This section presents comprehensive results for the \textbf{REST} benchmarks, including performance metrics across all evaluation conditions and detailed breakdowns by modality.

\subsection{REST}
\paragraph{Cross-Modal Consistency Analysis}
Tables \ref{tab:app_acc_ocr} and \ref{tab:app_acc_all} present RER and CFR scores for the OCR-correct subset and the complete set, respectively. Results are given per benchmark (MMLU, ARC, GSM8k-Symbolic, \textsc{SoEBench}) with corresponding OCR accuracy rates. Notably, DeepSeek-Tiny achieves 0\% OCR accuracy on \textsc{SoEBench}, resulting in worst-case consistency scores for this model-benchmark combination. The minimal difference between OCR-correct and full dataset scores validates our approach of constraining OCR complexity: cross-modal inconsistency persists even when text recognition is perfect, confirming that the phenomenon stems from fundamental reasoning differences rather than recognition failures.

\begin{table*}[htb]
\centering
\footnotesize 
\caption{\textbf{REST Consistency performance across VLMs when OCR Correct.} Columns show Render-Equivalence Rate (RER), Cross-Modality Failure Rate (CFR), and perfect OCR rate (OCR) for each benchmark component. On the left, scores for \textbf{REST}, which are the mean over all RER and CFR scores. \textbf{Results include only questions with correct OCR.}}
\setlength{\tabcolsep}{0.45em}
\begin{tabular}{l|cc|ccc|ccc|ccc|ccc}
\toprule
\textbf{Model} 
& \multicolumn{2}{c|}{\textbf{REST (Avg.)}} 
& \multicolumn{3}{c|}{\textbf{MMLU~\cite{hendrycks2020mmlu}}} 
& \multicolumn{3}{c|}{\textbf{AI2-ARC~\cite{clark2018arc}}} 
& \multicolumn{3}{c|}{\textbf{GSM-Sym~\cite{mirzadeh2024gsm}}} 
& \multicolumn{3}{c}{\textbf{SoEBench}} \\
\cmidrule(lr){2-3} \cmidrule(lr){4-6} \cmidrule(lr){7-9} \cmidrule(lr){10-12} \cmidrule(lr){13-15}
 & RER & CFR & RER & CFR & OCR & RER & CFR  & OCR & RER & CFR  & OCR & RER & CFR  & OCR \\
\midrule
Deepseek-Tiny & 6.6 & 98.0 & 2.3 & 97.8 & 100.0 & 4.0 & 95.6 & 100.0 & 20.3 & 98.7 & 100.0 & 0.0 & 100.0 & 100.0 \\
Phi-4 & 14.9 & 82.3 & 8.8 & 89.8 & 100.0 & 12.1 & 86.7 & 100.0 & 37.5 & 57.3 & 100.0 & 1.3 & 95.5 & 100.0 \\
Phi-3.5 & 19.4 & 79.3 & 21.5 & 80.3 & 100.0 & 28.4 & 70.7 & 100.0 & 27.7 & 66.2 & 100.0 & 0.0 & 100.0 & 100.0 \\
InternVL3 (2B) & 32.9 & 63.7 & 46.1 & 52.6 & 100.0 & 62.0 & 37.0 & 100.0 & 19.0 & 76.4 & 100.0 & 4.7 & 88.9 & 100.0 \\
Deepseek-Small & 35.9 & 60.9 & 42.6 & 56.9 & 100.0 & 57.8 & 40.6 & 100.0 & 42.7 & 50.1 & 100.0 & 0.7 & 95.8 & 100.0 \\
Gemma-3 (4B) & 53.9 & 42.3 & 53.3 & 46.1 & 100.0 & 65.0 & 33.8 & 100.0 & 55.4 & 38.7 & 100.0 & 41.9 & 50.5 & 100.0 \\
Gemini-2.5-flash-lite & 54.3 & 40.3 & 74.9 & 19.5 & 100.0 & 87.6 & 8.6 & 100.0 & 41.4 & 51.2 & 100.0 & 13.3 & 81.8 & 100.0 \\
Qwen-2.5 (7B) & 64.6 & 31.7 & 73.8 & 25.0 & 100.0 & 86.2 & 13.8 & 100.0 & 73.8 & 22.7 & 100.0 & 24.7 & 65.4 & 100.0 \\
GPT-4o-mini & 71.3 & 26.0 & 74.5 & 24.6 & 100.0 & 89.1 & 10.8 & 100.0 & 82.4 & 15.1 & 100.0 & 39.3 & 53.6 & 100.0 \\
Mistral-Small & 73.6 & 23.9 & 75.8 & 23.0 & 100.0 & 88.8 & 10.7 & 100.0 & 88.5 & 9.9 & 100.0 & 41.3 & 52.0 & 100.0 \\
Gemma-3 (12B) & 75.8 & 21.3 & 69.8 & 28.6 & 100.0 & 87.7 & 11.4 & 100.0 & 82.4 & 14.7 & 100.0 & 63.3 & 30.7 & 100.0 \\
InternVL3 (14B) & 78.4 & 19.6 & 81.7 & 17.8 & 100.0 & 93.8 & 5.6 & 100.0 & 87.5 & 11.5 & 100.0 & 50.7 & 43.6 & 100.0 \\
Qwen-2.5 (32B) & 84.7 & 13.6 & 83.3 & 15.1 & 100.0 & 92.9 & 5.3 & 100.0 & 93.4 & 5.8 & 100.0 & 69.1 & 28.0 & 100.0 \\
Haiku-4.5 (Claude) & 90.3 & 8.9 & 84.5 & 14.9 & 100.0 & 92.2 & 6.5 & 100.0 & 92.6 & 6.9 & 100.0 & 92.0 & 7.4 & 100.0 \\
GPT-5-mini & 90.7 & 8.7 & 85.8 & 13.3 & 100.0 & 93.3 & 6.4 & 100.0 & 91.9 & 7.0 & 100.0 & 91.9 & 8.1 & 100.0 \\
\bottomrule
\end{tabular}
\label{tab:app_acc_ocr}
\end{table*}

\begin{table*}[htb]
\centering
\footnotesize 
\caption{\textbf{REST Consistency performance across VLMs on all questions.} Columns show Render-Equivalence Rate (RER), Cross-Modality Failure Rate (CFR), and perfect OCR rate (OCR) for each benchmark component. On the left, scores for \textbf{REST}, which are the mean over all RER and CFR scores. \textbf{Results include all questions (including OCR incorrect).}}
\setlength{\tabcolsep}{0.45em}
\begin{tabular}{l|cc|ccc|ccc|ccc|ccc}
\toprule
\textbf{Model} 
& \multicolumn{2}{c|}{\textbf{REST (Avg.)}} 
& \multicolumn{3}{c|}{\textbf{MMLU~\cite{hendrycks2020mmlu}}} 
& \multicolumn{3}{c|}{\textbf{AI2-ARC~\cite{clark2018arc}}} 
& \multicolumn{3}{c|}{\textbf{GSM-Sym~\cite{mirzadeh2024gsm}}} 
& \multicolumn{3}{c}{\textbf{SoEBench}} \\
\cmidrule(lr){2-3} \cmidrule(lr){4-6} \cmidrule(lr){7-9} \cmidrule(lr){10-12} \cmidrule(lr){13-15}
 & RER & CFR & RER & CFR & OCR & RER & CFR  & OCR & RER & CFR  & OCR & RER & CFR  & OCR \\
\midrule
Deepseek-Tiny & 6.5 & 98.1 & 2.2 & 97.9 & 90.6 & 3.9 & 95.8 & 96.8 & 19.9 & 98.7 & 93.6 & 0.0 & 100.0 & 0.0 \\
Phi-4 & 14.5 & 82.8 & 7.8 & 90.9 & 84.9 & 12.1 & 86.9 & 97.5 & 36.7 & 58.0 & 93.5 & 1.3 & 95.5 & 100.0 \\
Phi-3.5 & 19.3 & 79.5 & 22.0 & 80.4 & 79.2 & 28.2 & 70.9 & 91.4 & 26.9 & 66.7 & 90.5 & 0.0 & 100.0 & 100.0 \\
InternVL3 (2B) & 32.6 & 63.9 & 45.3 & 53.5 & 88.6 & 61.6 & 37.2 & 96.6 & 18.7 & 76.1 & 77.5 & 4.7 & 88.9 & 100.0 \\
Deepseek-Small & 35.9 & 60.9 & 42.4 & 57.1 & 95.3 & 57.8 & 40.6 & 99.3 & 42.6 & 50.1 & 96.9 & 0.7 & 95.8 & 100.0 \\
Gemma-3 (4B) & 52.3 & 44.0 & 53.6 & 47.3 & 73.4 & 63.5 & 35.1 & 90.2 & 53.4 & 40.3 & 84.2 & 38.7 & 53.2 & 86.0 \\
Gemini-2.5-flash-lite & 54.1 & 40.4 & 74.4 & 19.8 & 96.9 & 87.7 & 8.5 & 99.6 & 41.1 & 51.2 & 96.9 & 13.3 & 81.8 & 100.0 \\
Qwen-2.5 (7B) & 64.3 & 31.9 & 73.4 & 25.3 & 97.9 & 86.2 & 13.8 & 99.8 & 72.9 & 22.9 & 97.5 & 24.7 & 65.4 & 100.0 \\
GPT-4o-mini & 71.2 & 26.1 & 74.2 & 24.8 & 97.5 & 89.0 & 10.9 & 99.6 & 82.2 & 15.3 & 98.2 & 39.3 & 53.6 & 100.0 \\
Mistral-Small & 73.4 & 24.1 & 75.6 & 23.3 & 97.7 & 88.8 & 10.7 & 99.9 & 88.1 & 10.4 & 95.9 & 41.3 & 52.0 & 100.0 \\
Gemma-3 (12B) & 75.5 & 21.6 & 69.5 & 28.9 & 93.6 & 87.7 & 11.4 & 98.9 & 81.7 & 15.4 & 93.5 & 63.3 & 30.7 & 100.0 \\
InternVL3 (14B) & 78.1 & 19.9 & 81.1 & 18.2 & 90.6 & 93.8 & 5.5 & 95.5 & 86.9 & 12.1 & 95.2 & 50.7 & 43.6 & 100.0 \\
Qwen-2.5 (32B) & 84.5 & 13.7 & 83.2 & 15.3 & 96.7 & 92.9 & 5.3 & 99.5 & 93.1 & 6.1 & 94.5 & 68.7 & 28.0 & 99.3 \\
Haiku-4.5 (Claude) & 90.1 & 9.1 & 84.3 & 15.0 & 97.7 & 92.2 & 6.4 & 99.7 & 91.8 & 7.8 & 95.6 & 92.0 & 7.4 & 100.0 \\
GPT-5-mini & 90.7 & 8.8 & 85.7 & 13.4 & 98.5 & 93.3 & 6.4 & 99.7 & 91.6 & 7.2 & 98.4 & 92.0 & 8.0 & 99.3 \\
\bottomrule
\end{tabular}
\label{tab:app_acc_all}
\end{table*}

\paragraph{Modality-Specific Performance}
Tables \ref{tab:accuracy_quadrant_ocr} and \ref{tab:accuracy_quadrant_all} provide accuracy scores across text, image, and mixed modalities for both evaluation subsets. Character Error Rate (CER) metrics provide a quantitative assessment of OCR performance across models. We additionally report results for the ``OCR-first" strategy, which yields mixed outcomes, improving performance for some model-benchmark combinations while degrading others. 

Notably, Max Modal Coverage (MMC) analysis reveals that several models achieve near-perfect solvability when considering their best-performing modality: GPT-5-mini attains 100\% MMC on SOEBENCH, 96.0\% on ARC, and 97.3\% on GSM8k-Symbolic. However, this high coverage does not translate to consistent performance. The same model exhibits variation across modalities, highlighting the gap between potential and realised performance under different input formats.

\begin{table*}[htb]
\centering
\footnotesize 
\caption{\textbf{Modality-specific accuracies (Text, Image, Mixed) across REST benchmarks.} MMC indicates maximum modal coverage (percentage of questions solved in at least one modality). The `OCR First' metrics represent the accuracy when first prompting to OCR and then solving the question. \textbf{Results include only questions with correct OCR.}}
\setlength{\tabcolsep}{0.4em} %
\begin{tabular}{@{}cc@{}}
  \begin{tabular}{@{}lcccccc@{}}
    \toprule
    \multicolumn{7}{c}{\textbf{MMLU}} \\
    \cmidrule(lr){1-7}
    Model                      & Text    & Mixed   & Image & \makecell{OCR \\ First} & CER & MMC \\
    \midrule
InternVL3 (14B) & 82.9 & 81.4 & 83.0 & 83.0 & 0.0 & 90.0 \\
InternVL3 (2B) & 61.6 & 59.0 & 56.6 & 55.4 & 0.0 & 79.6 \\
Mistral-Small & 84.1 & 78.9 & 82.5 & 82.6 & 0.0 & 91.3 \\
Phi-3.5 & 61.1 & 41.5 & 32.5 & 39.3 & 0.0 & 75.6 \\
Phi-4 & 47.0 & 44.2 & 31.4 & 33.6 & 0.0 & 80.1 \\
Qwen-2.5 (32B) & 83.3 & 84.1 & 83.5 & 80.0 & 0.0 & 90.0 \\
Qwen-2.5 (7B) & 72.6 & 71.8 & 72.9 & 72.9 & 0.0 & 82.4 \\
Haiku-4.5 (Claude) & 90.1 & 87.3 & 85.1 & 85.2 & 0.0 & 93.7 \\
Deepseek-Small & 54.9 & 54.9 & 51.2 & 38.1 & 0.0 & 75.2 \\
Deepseek-Tiny & 29.5 & 26.6 & 25.3 & 17.6 & 0.0 & 68.7 \\
Gemini-2.5-flash-lite & 81.6 & 79.7 & 78.4 & 78.2 & 0.0 & 88.1 \\
Gemma-3 (12B) & 77.6 & 76.7 & 72.2 & 72.2 & 0.0 & 87.1 \\
Gemma-3 (4B) & 68.0 & 64.7 & 57.2 & 57.5 & 0.0 & 81.3 \\
GPT-4o-mini & 84.4 & 77.2 & 77.0 & 71.9 & 0.0 & 89.8 \\
GPT-5-mini & 89.3 & 86.5 & 87.8 & 89.0 & 0.0 & 93.4 \\

  \end{tabular}
  &
  \begin{tabular}{@{}lcccccc@{}}
    \toprule
    \multicolumn{7}{c}{\textbf{GSM8K-Symbolic}} \\
    \cmidrule(lr){1-7}
    Model                      & Text    & Mixed   & Image & \makecell{OCR \\ First} & CER & MMC \\
    \midrule
InternVL3 (14B) & 91.2 & 91.9 & 93.0 & 92.5 & 0.0 & 96.8 \\
InternVL3 (2B) & 52.1 & 40.3 & 53.1 & 64.6 & 0.0 & 76.8 \\
Mistral-Small & 92.6 & 91.7 & 93.3 & 93.2 & 0.0 & 96.5 \\
Phi-3.5 & 66.9 & 42.9 & 48.3 & 45.2 & 0.0 & 78.3 \\
Phi-4 & 76.5 & 55.8 & 54.5 & 60.7 & 0.0 & 85.9 \\
Qwen-2.5 (32B) & 93.6 & 93.1 & 93.0 & 93.4 & 0.0 & 95.7 \\
Qwen-2.5 (7B) & 84.4 & 82.0 & 82.1 & 83.0 & 0.0 & 91.9 \\
Haiku-4.5 (Claude) & 95.6 & 94.9 & 94.1 & 92.8 & 0.0 & 97.7 \\
Deepseek-Small & 67.3 & 54.7 & 60.9 & 57.5 & 0.0 & 80.4 \\
Deepseek-Tiny & 15.5 & 0.8 & 0.7 & 0.3 & 0.0 & 16.0 \\
Gemini-2.5-flash-lite & 79.4 & 60.1 & 50.5 & 59.5 & 0.0 & 84.8 \\
Gemma-3 (12B) & 91.3 & 87.7 & 88.1 & 89.5 & 0.0 & 95.1 \\
Gemma-3 (4B) & 82.2 & 70.5 & 67.4 & 70.0 & 0.0 & 89.6 \\
GPT-4o-mini & 91.4 & 86.7 & 89.0 & 88.6 & 0.0 & 95.2 \\
GPT-5-mini & 95.1 & 93.7 & 94.1 & 95.0 & 0.0 & 97.4 \\

  \end{tabular}
  \\[1em]
  \begin{tabular}{@{}lcccccc@{}}
    \toprule
    \multicolumn{7}{c}{\textsc{SoEBench}} \\
    \cmidrule(lr){1-7}
    Model                      & Text    & Mixed   & Image & \makecell{OCR \\ First} & CER & MMC \\
    \midrule
InternVL3 (14B) & 73.3 & 70.0 & 72.7 & 71.3 & 0.0 & 88.7 \\
InternVL3 (2B) & 14.0 & 14.7 & 33.3 & 9.3 & 0.0 & 42.0 \\
Mistral-Small & 60.7 & 62.0 & 62.0 & 63.3 & 0.0 & 82.0 \\
Phi-3.5 & 1.3 & 1.3 & 0.7 & 0.7 & 0.0 & 3.3 \\
Phi-4 & 13.3 & 13.3 & 17.3 & 13.3 & 0.0 & 29.3 \\
Qwen-2.5 (32B) & 87.2 & 87.9 & 75.2 & 85.2 & 0.0 & 96.0 \\
Qwen-2.5 (7B) & 48.0 & 46.0 & 48.7 & 36.7 & 0.0 & 71.3 \\
Haiku-4.5 (Claude) & 96.7 & 95.3 & 97.3 & 96.0 & 0.0 & 99.3 \\
Deepseek-Small & 8.0 & 7.3 & 4.7 & 3.3 & 0.0 & 16.0 \\
Deepseek-Tiny & 0.7 & 0.0 & 0.0 & 0.0 & 70.5 & 0.7 \\
Gemini-2.5-flash-lite & 42.7 & 34.7 & 49.3 & 44.0 & 0.0 & 73.3 \\
Gemma-3 (12B) & 84.0 & 72.7 & 76.0 & 70.7 & 0.0 & 91.3 \\
Gemma-3 (4B) & 75.2 & 58.9 & 57.4 & 55.8 & 0.0 & 84.5 \\
GPT-4o-mini & 65.3 & 61.3 & 62.0 & 64.7 & 0.0 & 83.3 \\
GPT-5-mini & 98.7 & 95.3 & 97.3 & 98.0 & 0.0 & 100.0 \\
\bottomrule
  \end{tabular}
  &
  \begin{tabular}{@{}lcccccc@{}}
    \toprule
    \multicolumn{7}{c}{\textbf{AI2-ARC}} \\
    \cmidrule(lr){1-7}
    Model                      & Text    & Mixed   & Image & \makecell{OCR \\ First} & CER & MMC \\
    \midrule
InternVL3 (14B) & 92.9 & 92.9 & 94.2 & 94.2 & 0.0 & 95.7 \\
InternVL3 (2B) & 76.0 & 74.0 & 74.0 & 72.8 & 0.0 & 89.7 \\
Mistral-Small & 92.7 & 90.0 & 92.6 & 92.0 & 0.0 & 95.8 \\
Phi-3.5 & 73.3 & 54.3 & 37.9 & 57.0 & 0.0 & 83.6 \\
Phi-4 & 59.9 & 48.5 & 35.0 & 44.1 & 0.0 & 86.1 \\
Qwen-2.5 (32B) & 92.9 & 92.9 & 92.9 & 90.4 & 0.0 & 95.2 \\
Qwen-2.5 (7B) & 86.1 & 87.0 & 88.3 & 88.9 & 0.0 & 93.1 \\
Haiku-4.5 (Claude) & 93.3 & 93.3 & 92.7 & 92.0 & 0.0 & 95.7 \\
Deepseek-Small & 69.1 & 68.7 & 70.0 & 57.2 & 0.0 & 85.3 \\
Deepseek-Tiny & 34.2 & 28.7 & 27.5 & 19.7 & 0.0 & 70.6 \\
Gemini-2.5-flash-lite & 90.4 & 89.0 & 88.8 & 89.0 & 0.0 & 93.2 \\
Gemma-3 (12B) & 90.5 & 90.2 & 88.6 & 89.6 & 0.0 & 94.4 \\
Gemma-3 (4B) & 79.2 & 78.2 & 71.4 & 72.7 & 0.0 & 90.5 \\
GPT-4o-mini & 93.6 & 89.3 & 89.9 & 87.8 & 0.0 & 95.4 \\
GPT-5-mini & 94.6 & 92.7 & 93.3 & 94.4 & 0.0 & 96.0 \\
\bottomrule
  \end{tabular}
\end{tabular}
\label{tab:accuracy_quadrant_ocr}
\end{table*}

\begin{table*}[htb]
\centering
\footnotesize 
\caption{\textbf{Modality-specific accuracies (Text, Image, Mixed) across REST benchmarks.} MMC indicates maximum modal coverage (percentage of questions solved in at least one modality). The `OCR First' metrics represent the accuracy when first prompting to OCR and then solving the question. \textbf{Results include all questions (including OCR incorrect).}}
\setlength{\tabcolsep}{0.4em} %
\begin{tabular}{@{}cc@{}}
  \begin{tabular}{@{}lcccccc@{}}
    \toprule
    \multicolumn{7}{c}{\textbf{MMLU}} \\
    \cmidrule(lr){1-7}
    Model                      & Text    & Mixed   & Image & \makecell{OCR \\ First} & CER & MMC \\
    \midrule
InternVL3 (14B) & 82.4 & 81.0 & 82.5 & 82.6 & 0.5 & 89.7 \\
InternVL3 (2B) & 61.2 & 58.5 & 56.0 & 55.0 & 1.1 & 79.4 \\
Mistral-Small & 83.9 & 78.7 & 82.3 & 82.3 & 0.0 & 91.3 \\
Phi-3.5 & 58.1 & 40.3 & 31.3 & 37.1 & 4.5 & 72.8 \\
Phi-4 & 45.0 & 43.2 & 30.6 & 32.5 & 0.6 & 79.1 \\
Qwen-2.5 (32B) & 83.3 & 84.1 & 83.3 & 79.7 & 0.1 & 90.0 \\
Qwen-2.5 (7B) & 72.4 & 71.5 & 72.7 & 72.7 & 0.0 & 82.3 \\
Haiku-4.5 (Claude) & 90.0 & 87.2 & 85.1 & 85.2 & 0.1 & 93.7 \\
Deepseek-Small & 54.8 & 54.6 & 51.1 & 37.7 & 0.2 & 75.1 \\
Deepseek-Tiny & 29.5 & 26.2 & 25.2 & 17.5 & 1.3 & 68.6 \\
Gemini-2.5-flash-lite & 81.3 & 79.4 & 78.0 & 77.6 & 0.1 & 87.9 \\
Gemma-3 (12B) & 77.4 & 76.5 & 71.9 & 71.8 & 0.2 & 87.0 \\
Gemma-3 (4B) & 64.0 & 60.2 & 53.5 & 53.8 & 3.6 & 76.8 \\
GPT-4o-mini & 84.2 & 77.0 & 76.7 & 71.7 & 0.0 & 89.6 \\
GPT-5-mini & 89.2 & 86.5 & 87.7 & 89.0 & 0.0 & 93.4 \\

  \end{tabular}
  &
  \begin{tabular}{@{}lcccccc@{}}
    \toprule
    \multicolumn{7}{c}{\textbf{GSM8K-Symbolic}} \\
    \cmidrule(lr){1-7}
    Model                      & Text    & Mixed   & Image & \makecell{OCR \\ First} & CER & MMC \\
    \midrule
InternVL3 (14B) & 90.6 & 91.4 & 92.6 & 92.2 & 0.1 & 96.6 \\
InternVL3 (2B) & 50.8 & 38.7 & 51.8 & 62.2 & 3.8 & 74.8 \\
Mistral-Small & 92.2 & 91.3 & 92.9 & 92.8 & 0.0 & 96.3 \\
Phi-3.5 & 65.5 & 41.9 & 46.8 & 43.7 & 0.7 & 76.9 \\
Phi-4 & 75.5 & 55.1 & 53.7 & 60.1 & 0.1 & 85.1 \\
Qwen-2.5 (32B) & 93.2 & 92.8 & 92.7 & 92.5 & 0.1 & 95.4 \\
Qwen-2.5 (7B) & 83.6 & 81.2 & 81.3 & 82.1 & 0.0 & 91.1 \\
Haiku-4.5 (Claude) & 95.4 & 94.4 & 93.6 & 92.4 & 0.1 & 97.7 \\
Deepseek-Small & 67.2 & 54.6 & 60.7 & 57.2 & 0.1 & 80.2 \\
Deepseek-Tiny & 15.2 & 0.8 & 0.7 & 0.3 & 0.4 & 15.8 \\
Gemini-2.5-flash-lite & 78.8 & 59.6 & 50.1 & 59.2 & 0.0 & 84.2 \\
Gemma-3 (12B) & 90.9 & 87.2 & 87.7 & 89.0 & 0.3 & 95.0 \\
Gemma-3 (4B) & 81.1 & 68.1 & 65.4 & 68.2 & 1.8 & 88.4 \\
GPT-4o-mini & 91.5 & 86.6 & 88.8 & 88.3 & 0.0 & 95.3 \\
GPT-5-mini & 94.9 & 93.5 & 93.8 & 94.5 & 0.0 & 97.3 \\

  \end{tabular}
  \\[1em]
  \begin{tabular}{@{}lcccccc@{}}
    \toprule
    \multicolumn{7}{c}{\textsc{SoEBench}} \\
    \cmidrule(lr){1-7}
    Model                      & Text    & Mixed   & Image & \makecell{OCR \\ First} & CER & MMC \\
    \midrule
InternVL3 (14B) & 73.3 & 70.0 & 72.7 & 71.3 & 0.0 & 88.7 \\
InternVL3 (2B) & 14.0 & 14.7 & 33.3 & 9.3 & 0.0 & 42.0 \\
Mistral-Small & 60.7 & 62.0 & 62.0 & 63.3 & 0.0 & 82.0 \\
Phi-3.5 & 1.3 & 1.3 & 0.7 & 0.7 & 0.0 & 3.3 \\
Phi-4 & 13.3 & 13.3 & 17.3 & 13.3 & 0.0 & 29.3 \\
Qwen-2.5 (32B) & 86.7 & 87.3 & 74.7 & 84.7 & 0.0 & 95.3 \\
Qwen-2.5 (7B) & 48.0 & 46.0 & 48.7 & 36.7 & 0.0 & 71.3 \\
Haiku-4.5 (Claude) & 96.7 & 95.3 & 97.3 & 96.0 & 0.0 & 99.3 \\
Deepseek-Small & 8.0 & 7.3 & 4.7 & 3.3 & 0.0 & 16.0 \\
Deepseek-Tiny & 0.7 & 0.0 & 0.0 & 0.0 & 70.5 & 0.7 \\
Gemini-2.5-flash-lite & 42.7 & 34.7 & 49.3 & 44.0 & 0.0 & 73.3 \\
Gemma-3 (12B) & 84.0 & 72.7 & 76.0 & 70.7 & 0.0 & 91.3 \\
Gemma-3 (4B) & 74.0 & 55.3 & 52.7 & 52.0 & 1.2 & 82.7 \\
GPT-4o-mini & 65.3 & 61.3 & 62.0 & 64.7 & 0.0 & 83.3 \\
GPT-5-mini & 98.7 & 95.3 & 97.3 & 98.0 & 0.2 & 100.0 \\
\bottomrule
  \end{tabular}
  &
  \begin{tabular}{@{}lcccccc@{}}
    \toprule
    \multicolumn{7}{c}{\textbf{AI2-ARC}} \\
    \cmidrule(lr){1-7}
    Model                      & Text    & Mixed   & Image & \makecell{OCR \\ First} & CER & MMC \\
    \midrule
InternVL3 (14B) & 93.0 & 93.2 & 94.3 & 94.3 & 0.2 & 95.8 \\
InternVL3 (2B) & 75.8 & 73.6 & 73.5 & 72.3 & 1.7 & 89.5 \\
Mistral-Small & 92.7 & 90.0 & 92.6 & 92.0 & 0.0 & 95.8 \\
Phi-3.5 & 74.0 & 54.7 & 37.6 & 55.5 & 5.3 & 84.2 \\
Phi-4 & 59.9 & 48.3 & 35.0 & 44.0 & 0.1 & 86.0 \\
Qwen-2.5 (32B) & 93.0 & 93.0 & 92.9 & 90.4 & 0.0 & 95.2 \\
Qwen-2.5 (7B) & 86.1 & 87.1 & 88.3 & 88.9 & 0.0 & 93.1 \\
Haiku-4.5 (Claude) & 93.3 & 93.3 & 92.7 & 92.0 & 0.0 & 95.7 \\
Deepseek-Small & 69.1 & 68.8 & 70.0 & 57.2 & 0.0 & 85.4 \\
Deepseek-Tiny & 34.6 & 28.9 & 27.1 & 19.6 & 0.5 & 70.6 \\
Gemini-2.5-flash-lite & 90.4 & 89.0 & 88.9 & 89.0 & 0.2 & 93.2 \\
Gemma-3 (12B) & 90.6 & 90.2 & 88.6 & 89.6 & 0.0 & 94.4 \\
Gemma-3 (4B) & 77.9 & 76.7 & 70.4 & 71.2 & 1.4 & 89.6 \\
GPT-4o-mini & 93.6 & 89.3 & 90.0 & 87.7 & 0.0 & 95.4 \\
GPT-5-mini & 94.6 & 92.7 & 93.3 & 94.4 & 0.0 & 96.0 \\
\bottomrule
  \end{tabular}
\end{tabular}
\label{tab:accuracy_quadrant_all}
\end{table*}

\paragraph{DeepSeek-OCR}
We also evaluate DeepSeek-OCR on our benchmarks. We observe that the model is not able to follow the instructions in our prompts. When we use their prescribed OCR prompt, \textit{\textless image\textgreater\ Free OCR}, we obtain a perfect OCR score on \textsc{SoEBench}. However, when using our OCR prompt, DeepSeek-OCR does not produce a single completely correct output. Moreover, for all reasoning questions, the model achieves a 0\% score, indicating an inability to properly follow the instructions. 

\begin{figure*}
    \centering
    \includegraphics[width=\textwidth]{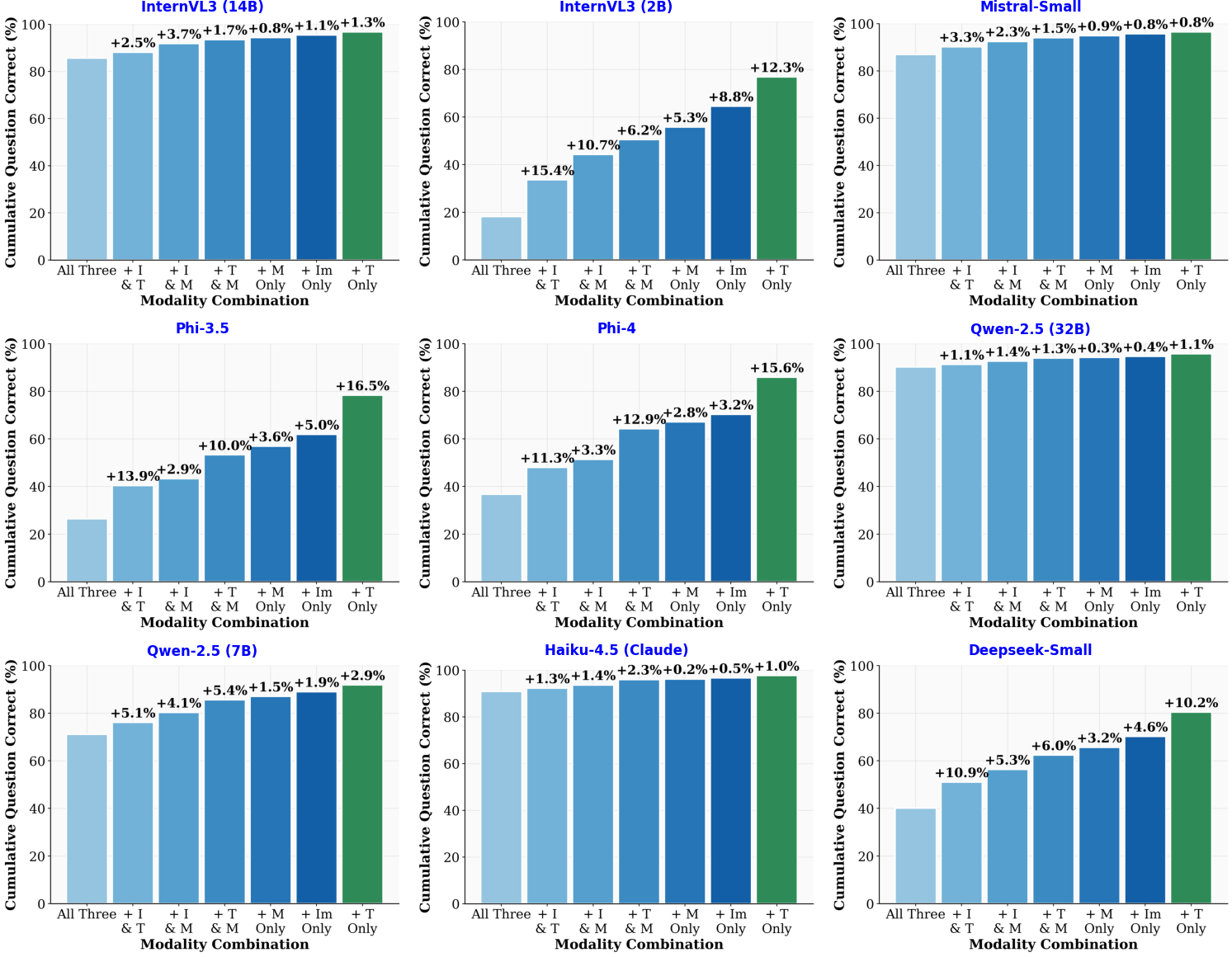}
    \caption{Cumulative distribution of correctly solved questions across modality combinations for models 1-8. Each step represents questions solvable in progressively fewer modalities, with the leftmost portion showing questions solved consistently across all three modalities.}
    \label{fig:stairs_case_part1}
\end{figure*}

\begin{figure*}
    \centering
    \includegraphics[width=\textwidth, trim=0 140 0 0, clip]{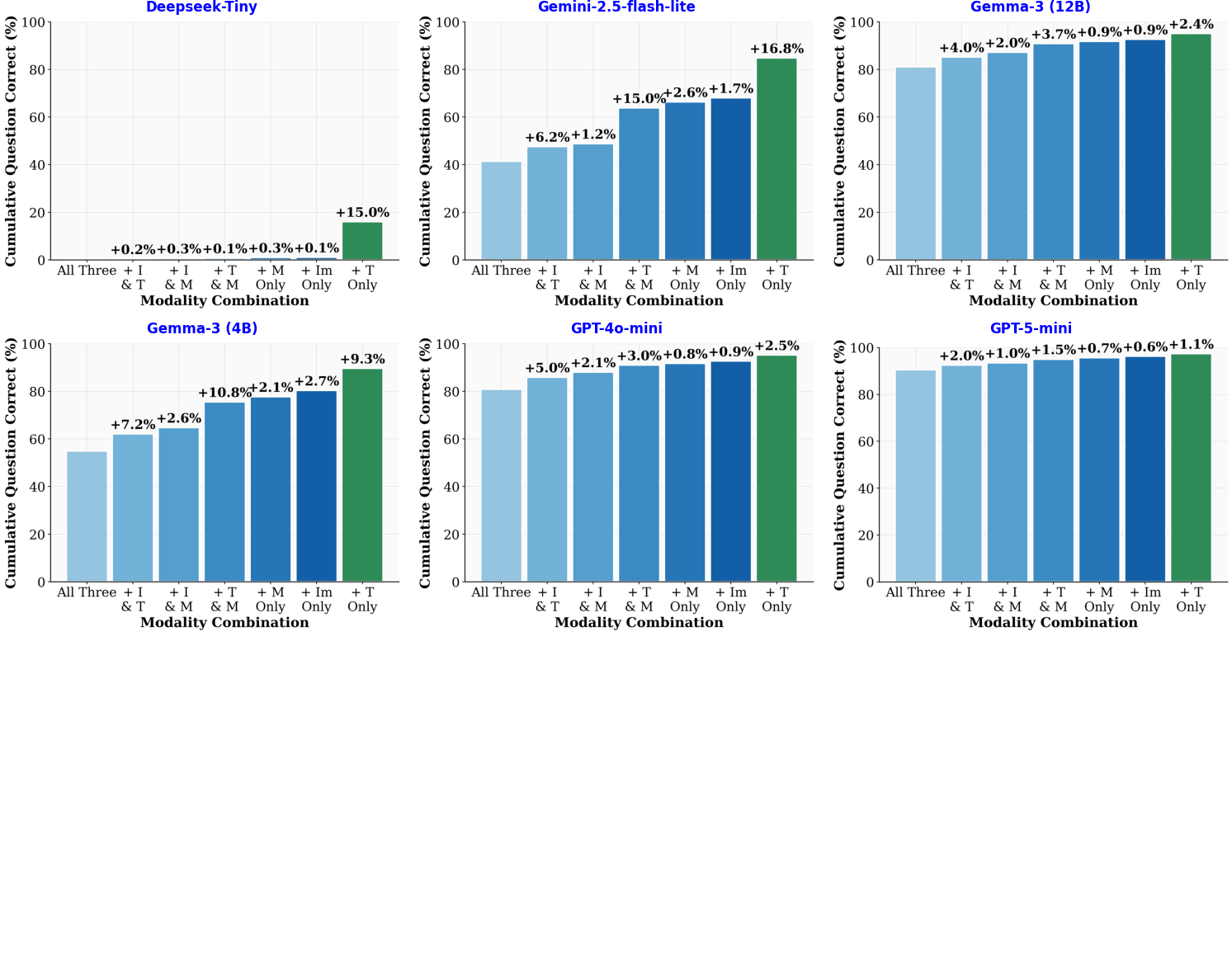}
    \caption{Cumulative distribution of correctly solved questions across modality combinations for models 9-15. Models with higher cross-modal consistency show larger proportions of questions solved in all modalities (leftmost region).}
    \label{fig:stairs_case_part2}
\end{figure*}

\paragraph{Distribution of Solvable Questions}
Figures \ref{fig:stairs_case_part1} and \ref{fig:stairs_case_part2} present staircase plots visualizing the cumulative distribution of correctly solved questions across modality combinations. These plots reveal distinct patterns: models with high consistency (e.g., GPT-5-mini) show steep initial rises with most questions solved across all modalities, while less consistent models exhibit gradual staircases indicating substantial modality-dependent performance variation.

\paragraph{Larger Models}
We evaluate InternVL-78B and Qwen-72B on REST (Table \ref{tab:app_reb_table1}). Despite increased capacity, GPT-5-mini ($\sim$ \href{https://www.r-bloggers.com/2025/08/how-many-parameters-does-gpt-5-have/}{27-149B}) still achieves the highest consistency scores, and inconsistency decreases only marginally; suggesting this is not purely a scaling issue. Due to resource constraints, we were not able to run bigger models on more data.

\begin{table}[h!]
\centering
\footnotesize 
\caption{GPT-5-mini shows the least inconsistency.}
\vspace{-0.25em}
\setlength{\tabcolsep}{0.45em}
\begin{tabular}{l|cc|ccc}
\toprule
\textbf{Model} 
& \multicolumn{2}{c}{\textbf{REST} (OCR\checkmark) }
& \multicolumn{3}{c}{\textbf{REST} (All Questions) } \\
\cmidrule(lr){2-3} \cmidrule(lr){4-6}
 & RER $\uparrow$ & CFR $\downarrow$ & RER $\uparrow$ & CFR $\downarrow$ & OCR\checkmark $\uparrow$  \\
 \midrule
InternVL3 (14B)&  78.4 & 19.6&  78.1 & 19.9&  95.3 \\
InternVL3 (78B) & 80.8 & 16.8 & 80.7 & 17.0 & 92.5 \\
Qwen-2.5 (72B) & 83.7 & 13.5 & 83.7 & 13.6 & \textbf{99.4} \\
GPT-5-mini & \textbf{90.7} & \textbf{8.7} & \textbf{90.7} &\textbf{ 8.8} & 99.0 \\
\bottomrule
\end{tabular}
\label{tab:app_reb_table1}
\end{table}

\paragraph{Mixed Design}
The mixed modality evaluates cross-modal integration, and is not designed to be a neutral midpoint between text and image. To test whether either configuration is inherently easier, we flip the setup: swapping text and image components. Results (Table \ref{tab:app_reb_mixed}) show 1--5\% differences, with no consistently easier design across MLLMs.
\begin{table}[h!!]
\centering
\footnotesize 
\caption{Flipping modality changes scores slightly (OCR correct).
}
\vspace{-0.25em}
\setlength{\tabcolsep}{0.35em}
\begin{tabular}{l|cc|cc|cc}
\toprule
\textbf{Model} 
& \multicolumn{2}{c|}{\textbf{AI2-ARC~} } 
& \multicolumn{2}{c|}{\textbf{GSM-Sym} } 
& \multicolumn{2}{c}{\textbf{MMLU} } \\
\cmidrule(lr){2-3} \cmidrule(lr){4-5} \cmidrule(lr){6-7}
 \textbf{Accuracy mixed $\uparrow$} & Orig & Flip & Orig & Flip & Orig & Flip \\
 \midrule
Qwen2.5 (7B) & \textbf{87.0} & 81.3 & \textbf{82.0} & 78.9 & \textbf{71.8} & 67.3 \\
Gemma (12B) & \textbf{90.2} & 86.0 & 87.7 & \textbf{87.9} & \textbf{76.7} & 74.1 \\
InternVL3 (14B) & \textbf{92.9} & 91.3 & \textbf{91.9} & 89.2 & \textbf{81.4} & 76.8 \\
GPT-5-mini & 92.7 & \textbf{93.9} & 93.7 & \textbf{94.2} & 86.5 & \textbf{88.6} \\
\bottomrule
\end{tabular}
\label{tab:app_reb_mixed}
\end{table}

\clearpage

\subsection{REST+}
\paragraph{Cross-modal Consistency}
Tables \ref{tab:rest_plus_core_ocr} and \ref{tab:rest_plus_core_oall} present \textbf{REST+} performance metrics across all visual permutations. We report RER consistency scores, modality-specific accuracies, and OCR success rates for both the OCR-correct subset and the complete dataset. Image accuracy represents the mean performance across all 10 visual permutations per question. Consistent with \textbf{REST} findings, text modality systematically outperforms image modality across all models. The reduced OCR accuracy demonstrates that lower resolution substantially impacts both text recognition and downstream reasoning.
\begin{table}[htb]
\centering
\footnotesize 
\caption{\textbf{REST+ performance on OCR-correct subset.} Render Equivalence Rate (RER) and modality-specific accuracies for questions where text recognition was perfect. Image accuracy is averaged across all visual permutations (3 fonts × 3 resolutions + colour variant) if OCR was correct.}
\setlength{\tabcolsep}{0.45em}
\begin{tabular}{l|ccccccc|}
\toprule
\textbf{Model} 
 & RER & Text & Image \\
\midrule
Deepseek-Tiny & 5.8 & 28.6 & 24.4 \\
Phi-4 & 11.3 & 43.9 & 32.7  \\
Phi-3.5 & 14.1 & 57.6 & 32.6 \\
Deepseek-Small & 24.5 & 55.5 & 51.9  \\
InternVL3 (2B) & 31.8 & 62.9 & 57.3 \\
Gemma-3 (4B) & 36.8 & 63.5 & 58.2  \\
Gemma-3 (12B) & 53.2 & 77.2 & 73.3  \\
Qwen-2.5 (7B) & 53.9 & 70.0 & 69.8 \\
GPT-4o-mini & 61.2 & 83.9 & 77.6 \\
Mistral-Small & 61.6 & 81.9 & 80.9 \\
Gemini-2.5-flash-lite & 63.8 & 82.7 & 77.3 \\
Haiku-4.5 (Claude) & 65.7 & 85.5 & 83.7  \\
GPT-5-mini & 67.6 & 89.4 & 86.3 \\
Qwen-2.5 (32B) & 69.4 & 80.0 & 82.0 \\
InternVL3 (14B) & 72.1 & 82.1 & 82.6 \\
\bottomrule
\end{tabular}
\label{tab:rest_plus_core_ocr}
\end{table}

\begin{table}[htb]
\centering
\footnotesize 
\caption{\textbf{REST+ performance on complete dataset.} Comprehensive evaluation, including all questions regardless of OCR success. Lower RER scores compared to the OCR-correct subset reflect the compounding effects of recognition errors and reasoning inconsistencies.}
\label{tab:rest_plus_core_all}\setlength{\tabcolsep}{0.45em}
\begin{tabular}{l|cccccccc|}
\toprule
\textbf{Model} 
 & RER & Text & Image & OCR \\
\midrule
Deepseek-Tiny & 3.5 & 28.6 & 24.8 & 82.7 \\
Phi-3.5 & 7.6 & 57.6 & 31.0 & 77.3 \\
Phi-4 & 7.6 & 43.9 & 32.4 & 80.1 \\
Deepseek-Small & 21.0 & 55.5 & 51.0 & 91.5 \\
InternVL3 (2B) & 27.5 & 62.9 & 55.5 & 85.2 \\
Gemma-3 (4B) & 27.6 & 63.5 & 53.5 & 70.5 \\
Haiku-4.5 (Claude) & 45.8 & 85.5 & 80.5 & 82.6 \\
Qwen-2.5 (7B) & 49.7 & 70.0 & 68.9 & 91.7 \\
Gemma-3 (12B) & 49.9 & 77.2 & 71.7 & 86.6 \\
Mistral-Small & 55.4 & 81.9 & 79.5 & 89.6 \\
GPT-4o-mini & 60.8 & 83.9 & 77.4 & 95.7 \\
Gemini-2.5-flash-lite & 62.7 & 82.7 & 77.0 & 95.9 \\
GPT-5-mini & 64.8 & 89.4 & 85.9 & 94.5 \\
Qwen-2.5 (32B) & 65.5 & 80.0 & 81.3 & 90.7 \\
InternVL3 (14B) & 69.3 & 82.1 & 82.0 & 88.8 \\
\bottomrule
\end{tabular}
\label{tab:rest_plus_core_oall}
\end{table}

\paragraph{Influence of DPI}
Tables \ref{tab:app_res_ocr} and \ref{tab:app_res_all} present resolution-dependent performance analysis for \textbf{REST+}, showing image accuracy and RER consistency scores across three DPI levels (50, 100, 200). Results reveal distinct model robustness patterns: while some models (DeepSeek-Small, Mistral-Small) exhibit performance degradation at lower resolutions, others (Qwen2.5-32B, InternVL3-14B) maintain stable performance across the different DPI levels.
\begin{table}[htb]
\centering
\footnotesize
\setlength{\tabcolsep}{0.45em}
\caption{\textbf{Image performance on REST+ for different DPI levels(OCR-correct subset).} Image accuracy and RER consistency scores stratified by resolution (50, 100, 200 DPI) for questions with perfect text recognition.}

\begin{tabular}{l|ccc|ccc|}
\toprule
\textbf{Model} 
& \multicolumn{3}{c|}{\textbf{Image Accuracy}} 
& \multicolumn{3}{c|}{\textbf{RER}} 
\\
\cmidrule(lr){2-4} \cmidrule(lr){5-7} 
\textbf{DPI} & 50 & 100 & 200 & 50 & 100 & 200
\\
\midrule
Deepseek-Small & 51.6 & 52.4 & 51.6 & 41.0 & 40.0 & 41.4 \\
Deepseek-Tiny & 23.9 & 24.7 & 24.3 & 14.9 & 12.8 & 10.0 \\
GPT-4o-mini & 78.0 & 78.7 & 76.1 & 72.6 & 73.7 & 72.8 \\
GPT-5-mini & 84.0 & 87.4 & 87.5 & 75.1 & 83.4 & 84.0 \\
Gemini-2.5-flash-lite & 78.1 & 77.3 & 77.0 & 71.5 & 71.0 & 71.7 \\
Gemma-3 (12B) & 76.1 & 73.4 & 71.4 & 73.0 & 67.3 & 65.7 \\
Gemma-3 (4B) & 58.2 & 57.4 & 58.8 & 59.3 & 54.4 & 54.6 \\
Haiku-4.5 (Claude) & 82.1 & 84.4 & 83.6 & 78.5 & 77.3 & 76.8 \\
InternVL3 (14B) & 80.7 & 82.6 & 84.2 & 78.4 & 81.5 & 82.2 \\
InternVL3 (2B) & 56.4 & 57.8 & 58.2 & 45.7 & 46.2 & 46.9 \\
Mistral-Small & 78.3 & 82.0 & 81.5 & 69.7 & 77.8 & 79.0 \\
Phi-3.5 & 32.1 & 32.6 & 32.6 & 26.7 & 26.8 & 25.8 \\
Phi-4 & 34.4 & 34.8 & 30.0 & 21.7 & 18.0 & 14.1 \\
Qwen-2.5 (32B) & 81.8 & 81.8 & 81.9 & 77.9 & 78.7 & 79.8 \\
Qwen-2.5 (7B) & 67.3 & 69.3 & 71.9 & 65.8 & 67.4 & 71.4 \\
\bottomrule
\end{tabular}
\label{tab:app_res_ocr}
\end{table}

\begin{table}[htb]
\centering
\footnotesize
\setlength{\tabcolsep}{0.35em}
\caption{\textbf{Image performance on REST+ for different DPI levels}  (complete set of questions). Image accuracy, RER consistency and OCR correct scores stratified by resolution (50, 100, 200 DPI) for all questions.}
\begin{tabular}{l|ccc|ccc|ccc}
\toprule
\textbf{Model} 
& \multicolumn{3}{c|}{\textbf{Img Acc.}} 
& \multicolumn{3}{c|}{\textbf{RER}} 
& \multicolumn{3}{c}{\textbf{OCR}} 
\\
\cmidrule(lr){2-4} \cmidrule(lr){5-7} \cmidrule(lr){8-10}
\textbf{DPI} & 50 & 100 & 200 & 50 & 100 & 200 & 50 & 100 & 200
\\
\midrule
Deepseek-Small & 49.7 & 51.8 & 51.2 & 32.7 & 36.7 & 39.2 & 85.3 & 93.4 & 94.9 \\
Deepseek-Tiny & 24.5 & 24.7 & 24.6 & 6.6 & 7.3 & 6.0 & 72.1 & 85.8 & 88.4 \\
GPT-4o-mini & 77.6 & 78.6 & 76.1 & 70.8 & 73.2 & 72.2 & 93.5 & 96.4 & 96.9 \\
GPT-5-mini & 83.1 & 87.1 & 87.5 & 71.5 & 82.7 & 83.6 & 89.6 & 95.9 & 97.1 \\
Gemini-2.5-FL & 77.7 & 77.2 & 76.8 & 70.6 & 70.4 & 70.9 & 95.1 & 96.2 & 96.3 \\
Gemma-3 (12B) & 71.6 & 72.6 & 70.9 & 62.5 & 64.9 & 63.6 & 73.9 & 90.9 & 92.9 \\
Gemma-3 (4B) & 52.1 & 54.7 & 53.8 & 41.7 & 46.6 & 44.9 & 63.1 & 76.5 & 71.7 \\
Haiku-4.5  & 73.4 & 84.0 & 83.8 & 53.6 & 74.4 & 76.3 & 57.4 & 92.2 & 94.3 \\
InternVL3 (14B) & 80.2 & 82.1 & 83.5 & 75.3 & 79.0 & 79.9 & 88.9 & 89.1 & 88.6 \\
InternVL3 (2B) & 54.2 & 55.8 & 56.7 & 39.7 & 40.4 & 41.4 & 84.5 & 84.7 & 86.1 \\
Mistral-Small & 74.6 & 81.8 & 81.4 & 60.0 & 76.9 & 78.4 & 73.9 & 95.8 & 96.6 \\
Phi-3.5 & 30.2 & 30.9 & 31.4 & 18.0 & 16.5 & 17.1 & 75.7 & 76.0 & 79.6 \\
Phi-4 & 32.8 & 34.6 & 30.3 & 10.9 & 13.3 & 10.2 & 62.7 & 86.3 & 88.5 \\
Qwen-2.5 (32B) & 79.7 & 81.7 & 81.9 & 71.2 & 78.1 & 79.5 & 77.8 & 95.5 & 96.9 \\
Qwen-2.5 (7B) & 65.2 & 69.2 & 71.5 & 58.1 & 66.5 & 69.8 & 79.1 & 97.0 & 97.2 \\
\bottomrule
\end{tabular}

\label{tab:app_res_all}
\end{table}

\paragraph{Influence of Font Family}
Tables \ref{tab:font_ocr} and \ref{tab:font_all} present font-specific performance metrics across DejaVu Sans, Courier New, and Cursive font families. Most models demonstrate consistent performance across font families. Claude Haiku-4.5 shows a performance drop specifically for Courier New in the complete dataset (but not in the OCR-correct subset), suggesting that it struggles with OCR on monospace font characteristics rather than having inherent reasoning limitations.
\begin{table}[htb]
\centering
\footnotesize
\setlength{\tabcolsep}{0.45em}
\caption{\textbf{Image performance on REST+ for different font families (OCR-correct subset).} Image accuracy and RER consistency scores for different fonts for questions with perfect text recognition.}
\begin{tabular}{l|ccc|ccc|}
\toprule
\textbf{Model} 
& \multicolumn{3}{c|}{\textbf{Image Accuracy}} 
& \multicolumn{3}{c|}{\textbf{RER}} 
\\
\cmidrule(lr){2-4} \cmidrule(lr){5-7} 
\textbf{font} & \makecell{Deja. \\ Sans} & \makecell{Cour. \\ New} & \makecell{Curs.} & \makecell{Deja. \\ Sans} & \makecell{Cour. \\ New} & \makecell{Curs.}
\\
\midrule
Deepseek-Small & 53.6 & 49.3 & 52.6 & 42.8 & 40.2 & 42.3 \\
Deepseek-Tiny & 24.5 & 23.6 & 24.8 & 10.5 & 13.0 & 11.8 \\
GPT-4o-mini & 77.3 & 77.3 & 78.2 & 71.4 & 72.1 & 73.7 \\
GPT-5-mini & 86.9 & 85.2 & 87.0 & 81.1 & 76.4 & 82.0 \\
Gemini-2.5-flash-lite & 78.6 & 76.6 & 77.1 & 72.0 & 70.8 & 71.5 \\
Gemma-3 (12B) & 73.6 & 72.8 & 74.0 & 66.7 & 68.4 & 67.4 \\
Gemma-3 (4B) & 57.5 & 58.4 & 58.5 & 54.2 & 58.6 & 55.2 \\
Haiku-4.5 (Claude) & 84.3 & 83.2 & 83.1 & 76.3 & 78.6 & 77.2 \\
InternVL3 (14B) & 82.4 & 82.4 & 82.6 & 81.1 & 80.4 & 80.3 \\
InternVL3 (2B) & 57.0 & 56.1 & 59.4 & 46.7 & 44.1 & 53.0 \\
Mistral-Small & 80.2 & 81.1 & 81.1 & 71.6 & 77.0 & 74.2 \\
Phi-3.5 & 34.9 & 29.6 & 32.6 & 30.3 & 24.9 & 28.4 \\
Phi-4 & 33.9 & 32.4 & 32.4 & 17.1 & 16.1 & 17.0 \\
Qwen-2.5 (32B) & 81.8 & 82.4 & 81.4 & 77.8 & 80.1 & 78.7 \\
Qwen-2.5 (7B) & 69.6 & 69.1 & 70.1 & 65.6 & 67.5 & 68.0 \\
\bottomrule
\end{tabular}
\label{tab:font_ocr}
\end{table}

\begin{table}[htb]
\centering
\footnotesize
\setlength{\tabcolsep}{0.45em}
\caption{\textbf{Image performance on REST+ for different font families (complete set of questions).} Image accuracy and RER consistency scores for different fonts on all questions.}
\begin{tabular}{l|ccc|ccc|}
\toprule
\textbf{Model} 
& \multicolumn{3}{c|}{\textbf{Image Accuracy}} 
& \multicolumn{3}{c|}{\textbf{RER}} 
\\
\cmidrule(lr){2-4} \cmidrule(lr){5-7} 
\textbf{font} & \makecell{Deja. \\ Sans} & \makecell{Cour. \\ New} & \makecell{Curs.} & \makecell{Deja. \\ Sans} & \makecell{Cour. \\ New} & \makecell{Curs.}
\\
\midrule
Deepseek-Small & 52.8 & 47.7 & 52.2 & 39.5 & 34.3 & 38.7 \\
Deepseek-Tiny & 24.9 & 23.8 & 25.1 & 6.2 & 6.2 & 6.9 \\
GPT-4o-mini & 77.1 & 77.2 & 77.9 & 70.6 & 71.5 & 72.6 \\
GPT-5-mini & 86.7 & 84.7 & 86.3 & 80.2 & 73.6 & 79.8 \\
Gemini-2.5-flash-lite & 78.4 & 76.5 & 76.8 & 71.4 & 70.1 & 70.8 \\
Gemma-3 (12B) & 72.1 & 70.6 & 72.4 & 63.0 & 61.3 & 63.1 \\
Gemma-3 (4B) & 53.6 & 53.2 & 53.8 & 44.0 & 43.0 & 44.6 \\
Haiku-4.5 (Claude) & 82.7 & 76.3 & 82.2 & 74.7 & 54.9 & 74.2 \\
InternVL3 (14B) & 82.0 & 81.7 & 82.0 & 78.8 & 77.1 & 78.5 \\
InternVL3 (2B) & 55.9 & 54.6 & 56.1 & 42.2 & 37.7 & 41.1 \\
Mistral-Small & 79.8 & 77.2 & 80.8 & 69.7 & 63.4 & 72.8 \\
Phi-3.5 & 32.9 & 28.8 & 30.8 & 22.6 & 13.7 & 19.5 \\
Phi-4 & 33.6 & 31.5 & 32.5 & 10.9 & 9.2 & 10.8 \\
Qwen-2.5 (32B) & 81.1 & 80.9 & 81.3 & 75.7 & 73.8 & 77.8 \\
Qwen-2.5 (7B) & 69.0 & 67.2 & 69.7 & 63.0 & 60.1 & 66.3 \\
\bottomrule
\end{tabular}
\label{tab:font_all}
\end{table}

\paragraph{Influence of colour}
Similarly, Tables \ref{tab:colour_ocr} and \ref{tab:colour_all} analyse the effect of text colour on image modality accuracy. We report performance for black text across all resolutions and specifically at 200 DPI to enable fair comparison with colored variants (all rendered at 200 DPI). Remarkably, every evaluated model achieves higher accuracy with at least one colour compared to black text at equivalent resolution. 
\begin{table}[htb]
\centering
\footnotesize 
\caption{\textbf{Text color effects on REST+ image accuracy (OCR-correct subset).} Comparison of accuracy for black text at multiple resolutions (50, 100, 200 DPI) versus colored text variants (all at 200 DPI). Numbers indicate the percentage of correctly solved questions. \textbf{Results shown for OCR-correct subset only.}}
\setlength{\tabcolsep}{0.4em}
\begin{tabular}{l|c|ccccccc}
\toprule
\textbf{Model}
& \multicolumn{1}{c|}{\makecell{\textbf{All}\\ \textbf{DPI}}} 
& \multicolumn{7}{c}{\textbf{DPI@200}}
\\
\cmidrule(lr){2-2} \cmidrule(lr){3-9}
& \makecell{\textbf{Bl.} \\ {\color{black}\rule{4mm}{2mm}}}
& \makecell{\textbf{Bl.} \\ {\color{black}\rule{4mm}{2mm}}}
& \makecell{\textbf{R} \\ {\color{red}\rule{4mm}{2mm}}}
& \makecell{\textbf{G} \\ {\color{green}\rule{4mm}{2mm}}}
& \makecell{\textbf{B} \\ {\color{blue}\rule{4mm}{2mm}}}
& \makecell{\textbf{Y} \\ {\color{yellow}\rule{4mm}{2mm}}}
& \makecell{\textbf{M} \\ {\color{magenta}\rule{4mm}{2mm}}}
& \makecell{\textbf{C} \\ {\color{cyan}\rule{4mm}{2mm}}}
\\
\midrule
Deepseek-Small & 51.9 & 51.6 & 55.5 & 52.3 & 51.7 & 56.6 & 49.7 & 49.1 \\
Deepseek-Tiny & 24.3 & 24.3 & 25.0 & 28.5 & 23.7 & 24.0 & 25.5 & 25.9 \\
GPT-4o-mini & 77.6 & 76.1 & 79.8 & 76.4 & 73.9 & 78.9 & 79.0 & 75.1 \\
GPT-5-mini & 86.3 & 87.5 & 84.9 & 87.6 & 80.4 & 89.9 & 86.4 & 86.8 \\
Gemini-2.5 FL & 77.5 & 77.0 & 81.5 & 75.0 & 71.0 & 74.7 & 74.4 & 77.4 \\
Gemma-3 (12B) & 73.5 & 71.4 & 77.2 & 70.2 & 70.9 & 67.1 & 72.6 & 75.6 \\
Gemma-3 (4B) & 58.1 & 58.8 & 62.0 & 58.7 & 61.7 & 56.0 & 58.1 & 53.3 \\
Haiku-4.5 (Claude) & 83.6 & 83.6 & 86.5 & 86.3 & 79.8 & 86.8 & 84.1 & 85.3 \\
InternVL3 (14B) & 82.5 & 84.2 & 87.6 & 83.2 & 81.0 & 84.6 & 84.1 & 79.5 \\
InternVL3 (2B) & 57.5 & 58.2 & 59.4 & 51.9 & 60.1 & 56.6 & 54.4 & 53.2 \\
Mistral-Small & 80.8 & 81.5 & 83.6 & 81.5 & 80.5 & 83.8 & 82.9 & 78.7 \\
Phi-3.5 & 32.5 & 32.6 & 32.9 & 32.6 & 33.1 & 36.0 & 34.0 & 37.5 \\
Phi-4 & 32.9 & 30.0 & 26.3 & 32.0 & 30.5 & 34.6 & 32.0 & 33.1 \\
Qwen-2.5 (32B) & 81.9 & 81.9 & 86.8 & 82.5 & 80.5 & 85.5 & 81.0 & 83.2 \\
Qwen-2.5 (7B) & 69.6 & 71.9 & 74.6 & 67.8 & 68.2 & 71.8 & 78.4 & 66.7 \\

\bottomrule
\end{tabular}
\label{tab:colour_ocr}
\end{table}

\begin{table}[htb]
\centering
\footnotesize 
\caption{\textbf{Text color effects on REST+ image accuracy (Complete set).} Comparison of accuracy for black text at multiple resolutions (50, 100, 200 DPI) versus colored text variants (all at 200 DPI). Numbers indicate the percentage of correctly solved questions. \textbf{Results shown for the complete dataset.}}
\setlength{\tabcolsep}{0.4em}

\begin{tabular}{l|c|ccccccc}
\toprule
\textbf{Model}
& \multicolumn{1}{c|}{\makecell{\textbf{All}\\ \textbf{DPI}}} 
& \multicolumn{7}{c}{\textbf{DPI@200}}
\\
\cmidrule(lr){2-2} \cmidrule(lr){3-9}
& \makecell{\textbf{Bl.} \\ {\color{black}\rule{4mm}{2mm}}}
& \makecell{\textbf{Bl.} \\ {\color{black}\rule{4mm}{2mm}}}
& \makecell{\textbf{R} \\ {\color{red}\rule{4mm}{2mm}}}
& \makecell{\textbf{G} \\ {\color{green}\rule{4mm}{2mm}}}
& \makecell{\textbf{B} \\ {\color{blue}\rule{4mm}{2mm}}}
& \makecell{\textbf{Y} \\ {\color{yellow}\rule{4mm}{2mm}}}
& \makecell{\textbf{M} \\ {\color{magenta}\rule{4mm}{2mm}}}
& \makecell{\textbf{C} \\ {\color{cyan}\rule{4mm}{2mm}}}
\\
\midrule
Deepseek-Small & 50.9 & 51.2 & 54.7 & 51.4 & 51.9 & 56.9 & 49.4 & 48.3 \\
Deepseek-Tiny & 24.6 & 24.6 & 26.0 & 28.2 & 24.3 & 23.2 & 27.2 & 27.8 \\
GPT-4o-mini & 77.4 & 76.1 & 79.6 & 76.2 & 74.0 & 79.0 & 77.8 & 75.0 \\
GPT-5-mini & 85.9 & 87.5 & 85.1 & 87.3 & 80.7 & 90.1 & 86.7 & 86.7 \\
Gemini-2.5 FL & 77.2 & 76.8 & 81.8 & 74.0 & 71.3 & 73.5 & 73.9 & 77.8 \\
Gemma-3 (12B) & 71.7 & 70.9 & 76.2 & 70.7 & 70.2 & 68.5 & 71.7 & 75.0 \\
Gemma-3 (4B) & 53.5 & 53.8 & 57.5 & 50.8 & 54.1 & 50.8 & 56.1 & 52.2 \\
Haiku-4.5 (Claude) & 80.4 & 83.8 & 82.9 & 82.9 & 76.7 & 82.9 & 80.6 & 81.7 \\
InternVL3 (14B) & 81.9 & 83.5 & 86.2 & 82.9 & 80.1 & 84.5 & 84.4 & 80.6 \\
InternVL3 (2B) & 55.5 & 56.7 & 58.6 & 50.8 & 60.8 & 53.0 & 55.6 & 53.9 \\
Mistral-Small & 79.3 & 81.4 & 82.9 & 81.8 & 80.1 & 82.3 & 82.2 & 79.4 \\
Phi-3.5 & 30.8 & 31.4 & 32.0 & 30.4 & 32.0 & 32.0 & 33.9 & 36.7 \\
Phi-4 & 32.6 & 30.3 & 26.5 & 32.6 & 30.4 & 32.0 & 32.2 & 34.4 \\
Qwen-2.5 (32B) & 81.1 & 81.9 & 86.2 & 82.9 & 80.7 & 85.1 & 80.6 & 82.8 \\
Qwen-2.5 (7B) & 68.6 & 71.5 & 74.0 & 68.0 & 68.5 & 71.3 & 77.8 & 66.1 \\

\bottomrule
\end{tabular}
\label{tab:colour_all}
\end{table}

\paragraph{Token Usage}
Finally, we show the number of tokens used by MLLMs in text and at the different DPI levels in Table \ref{tab:token_usage}. As mentioned in the paper, we see that fewer text tokens are needed to achieve the same level of accuracy, or more vision tokens are necessary to achieve higher accuracy than text, with the exception of Qwen2.5-VL-32B
\begin{table}[htb]
\centering
\footnotesize 
\caption{\textbf{Token consumption across modalities in REST+.} Average number of tokens processed for text modality versus image modality at three resolutions (50, 100, 200 DPI). Vision token counts exclude instruction tokens and represent only image encoding.}

\setlength{\tabcolsep}{0.45em}

\begin{tabular}{l|c|ccc}
\toprule
\textbf{Model} 
& \textbf{Text} 
& \multicolumn{3}{c}{\textbf{Image}}
\\
\cmidrule(lr){3-5}
& All.
& \makecell{DPI \\ 50}
& \makecell{DPI \\ 100}
& \makecell{DPI \\ 200}
\\
\midrule
Deepseek-Small & 145 & 584 & 896 & 1668 \\
Deepseek-Tiny & 126 & 584 & 896 & 1668 \\
GPT-4o-mini & - & - & - & - \\
GPT-5-mini & - & - & - & - \\
Gemini-2.5-flash-lite & - & - & - & - \\
Gemma-3 (12B) & 141 & 256 & 698 & 1051 \\
Gemma-3 (4B) & 141 & 256 & 698 & 1051 \\
Haiku-4.5 (Claude) & - & - & - & - \\
InternVL3 (14B) & 168 & 1683 & 1616 & 1631 \\
InternVL3 (2B) & 168 & 1683 & 1616 & 1631 \\
Mistral-Small & 127 & 116 & 332 & 1011 \\
Phi-3.5 & 154 & 611 & 542 & 577 \\
Phi-4 & 122 & 425 & 706 & 1631 \\
Qwen-2.5 (32B) & 142 & 106 & 316 & 978 \\
Qwen-2.5 (7B) & 142 & 106 & 316 & 978 \\
\bottomrule
\end{tabular}

\label{tab:token_usage}
\end{table}

\paragraph{Correlation of REST with common benchmarks}
\begin{figure*}[htb]
     \includegraphics[width=0.99\textwidth]{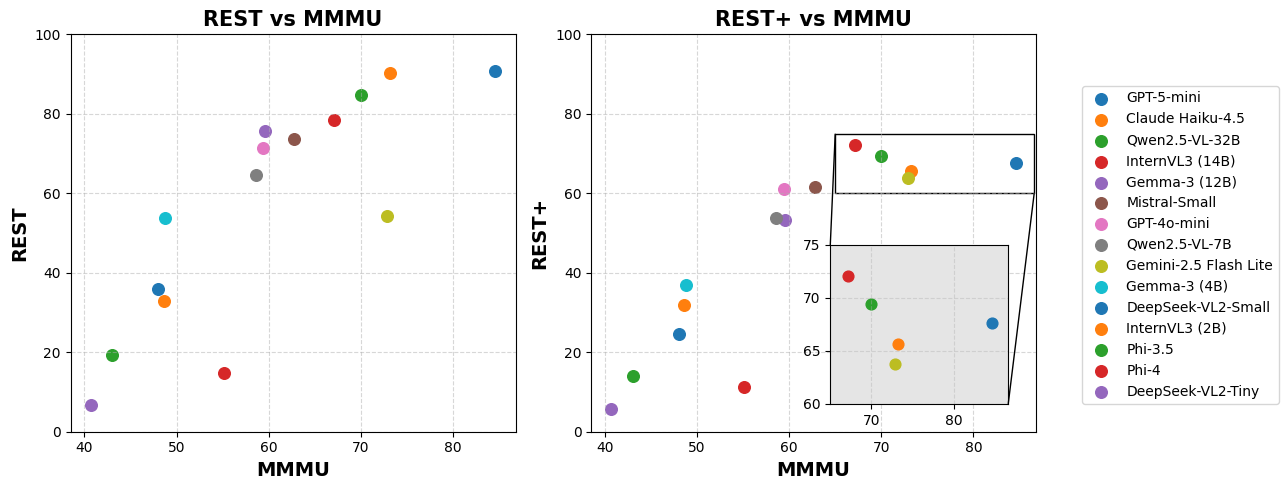}
    \captionof{figure}{Models that perform well on MMMU also score well on REST and REST+. The zoom inset shows that models with high MMMU scores do not generally obtain high REST+ scores.}
    \label{fig:mmmu}
\end{figure*}
We examine whether high-scoring MLLMs on general vision-language benchmarks show less cross-modal inconsistency. To this end, we plot the RER scores of \textbf{REST} and \textbf{REST+} against scores on MMMU \cite{yue2024mmmu}, see Figure \ref{fig:mmmu}. The MMMU scores were found on public benchmarks\footnote{\url{https://mmmu-benchmark.github.io/\#leaderboard}, \url{https://www.anthropic.com/news/claude-haiku-4-5}, \url{https://huggingface.co/Qwen/Qwen2.5-VL-32B-Instruct}, \url{https://huggingface.co/Qwen/Qwen2.5-VL-32B-Instruct}, \url{https://deepmind.google/models/gemini/flash-lite/}, \url{https://huggingface.co/OpenGVLab/InternVL3-2B}, \url{https://huggingface.co/microsoft/Phi-4-multimodal-instruct}, on Nov 20, 2025.}. Generally, we do find such a correlation, both for REST and REST+. Interestingly, the correlation seems low for models with high MMMU scores on REST+.

\subsection{Natural chess images instead of rendered text} \label{app:chess}

It is important to examine whether inconsistency extends beyond typographic inputs. We construct an additional same-content setting using chess positions, where spatial information translates naturally to text. Using ChessReD (Masouris 2023, \href{https://doi.org/10.4121/99b5c721-280b-450b-b058-b2900b69a90f}{DOI}), we ask yes-or-no questions about positions presented as (a) natural image, (b) generated image, or (c) text. 
\begin{figure}[tbh!]
    \begin{subfigure}[t]{0.26\columnwidth}
    \includegraphics[width=.9\linewidth]{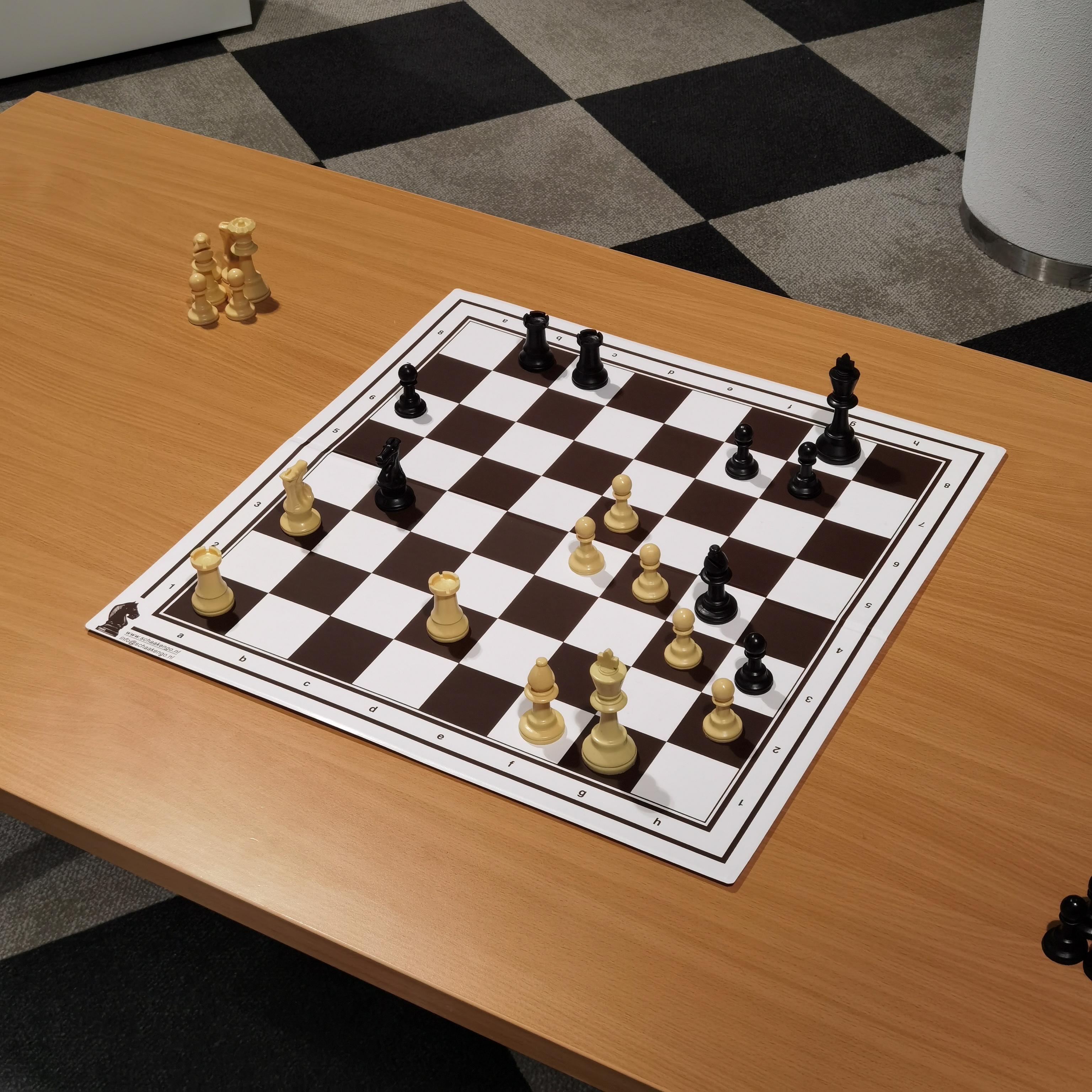}
    \caption{{Natural image (3D)}}
\end{subfigure}%
\hfill
\begin{subfigure}[t]{0.26\columnwidth}
    \includegraphics[width=.9\linewidth]{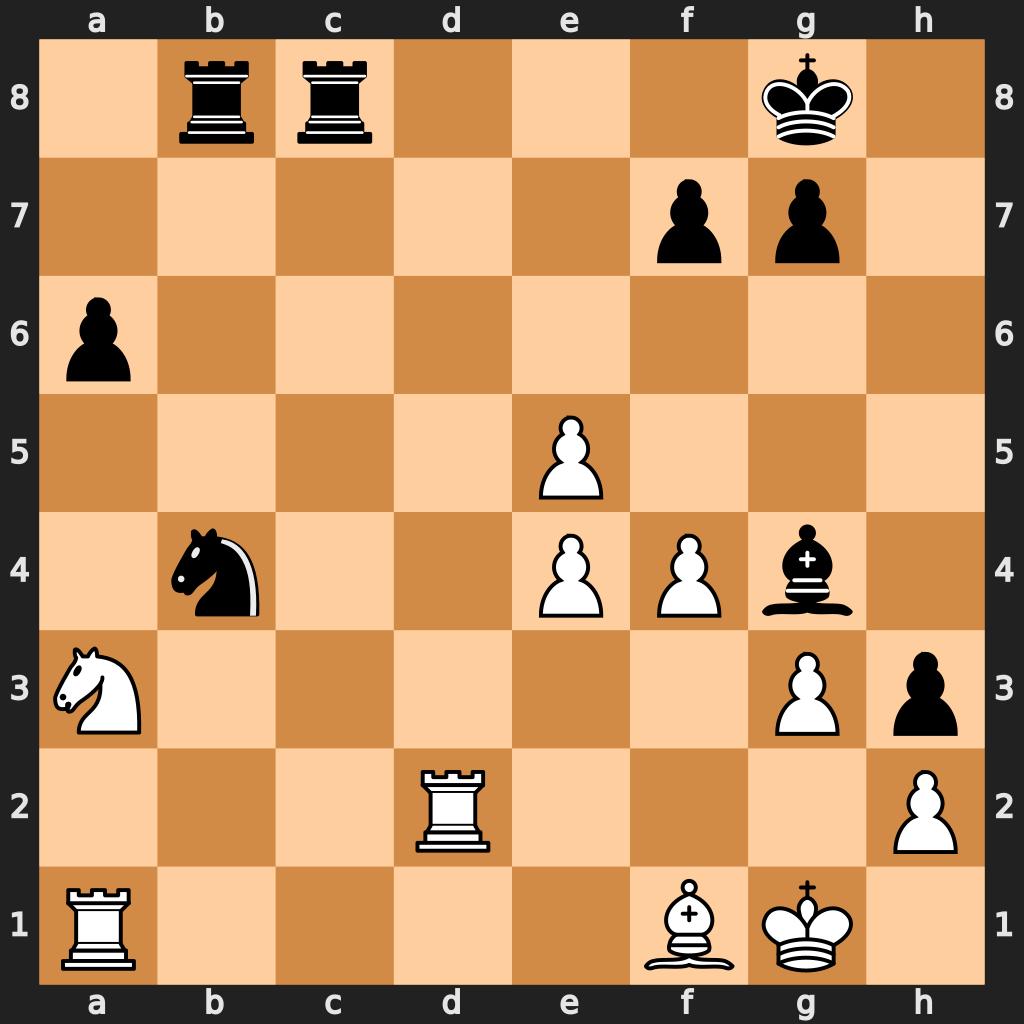}
    \caption{{Generated image (2D)}}
\end{subfigure}
\hfill
    \begin{subfigure}[t]{0.26\columnwidth}
        \includegraphics[width=\linewidth]{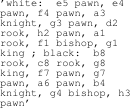}
        \caption{{Text}}
    \end{subfigure}
    \caption{\footnotesize Q1: Is a white pawn on the same row as another white pawn? (Y). Q2: Can the white bishop capture a black pawn in a single move? (Y).} \label{fig:chess}
\end{figure}
Results (Table \ref{tab:reb_chesstotal}) confirm that cross-modal inconsistency extends to natural images: answers vary by modality even when semantic understanding is verified (by querying the board state). Across modalities, the accuracy can differ max. 35\% (GPT-5-mini). Across models, RER ranges from 58.5 to 77.5, and models performing well on REST (GPT-5-mini, Mistral, Qwen-32B) also perform well on chess.
\begin{table}[tbh!]
\centering
\footnotesize 
\caption{{Cross-modal inconsistency for chess.} Sorted by RER.}
\setlength{\tabcolsep}{0.3em}
\begin{tabular}{l|ccc|cc|cc}
\toprule
\textbf{Model} 
& \multicolumn{5}{c|}{\makecell{\textbf{Chess} }} 
& \multicolumn{2}{c}{\makecell{\textbf{REST} }} \\
& 3D $\uparrow$ & 2D $\uparrow$ & Text $\uparrow$ & RER $\uparrow$ & CFR $\downarrow$ & RER $\uparrow$ & CFR $\downarrow$ \\
 \midrule
InternVL3 (14B) & 58.4 & 60.1 & 75.8 & 58.5 & 46.8 & 78.4 & 19.6 \\
Gemma-3 (4B) & 57.3 & 49.9 & 66.8 & 59.0 & 49.1 & 53.9 & 42.3 \\
Gemma-3 (12B) & 60.7 & 61.7 & 72.5 & 65.4 & 40.9 & 75.8 & 21.3 \\
Qwen2.5 (32B) & \textbf{64.9} & 69.4 & 85.3 & 65.5 & 35.7 & 84.7 & 13.6 \\
Mistral-Small & 62.6 & 74.2 & 75.7 & 67.7 & 36.7 & 73.6 & 23.9 \\
GPT-5-mini & 61.5 & \textbf{94.0 }& \textbf{96.4} & \textbf{77.5} & \textbf{22.9} & \textbf{90.7}  &\textbf{ 8.7} \\
\bottomrule
\end{tabular}
\label{tab:reb_chesstotal}
\end{table}

\end{document}